\documentclass{article}
\usepackage{nips13submit_e,times}
\usepackage{url}
\usepackage{color}
\usepackage{todonotes}

\usepackage{graphicx, subfigure}
\usepackage{amssymb}
\usepackage{epstopdf}
\usepackage{amsmath}
\usepackage{enumerate}
\usepackage[numbers]{natbib}

\newcommand{\reals}{\ensuremath{\mathbb{R}}}

\newcommand{\ux}{\ensuremath{\underline{x}}}
\newcommand{\ox}{\ensuremath{\overline{x}}}
\newcommand{\un}{\ensuremath{\underline{n}}}
\newcommand{\on}{\ensuremath{\overline{n}}}
\newcommand{\urho}{\ensuremath{\overleftarrow{\underline{\rho}}}}
\newcommand{\orho}{\ensuremath{\overleftarrow{\overline{\rho}}}}
\newcommand{\uv}{\ensuremath{\underline{v}}}
\newcommand{\ov}{\ensuremath{\overline{v}}}
\newcommand{\uu}{\ensuremath{\underline{u}}}
\newcommand{\ou}{\ensuremath{\overline{u}}}

\newcommand{\n}{\ensuremath{\eta}}

\title{A message-passing algorithm\\ for multi-agent trajectory planning}
\author{
Jos\' e Bento
\thanks{This author would like to thank Emily Hupf and Noa Ghersin for their support while writing this paper.}
\\
\texttt{jbento@disneyresearch.com} \\
\And
Nate Derbinsky\\
\texttt{nate.derbinsky@disneyresearch.com} \\
\And
Javier Alonso-Mora\\
\texttt{jalonso@disneyresearch.com} \\
\And
Jonathan Yedidia\\
\texttt{yedidia@disneyresearch.com}
}

\nipsfinalcopy 

\begin{document}
\maketitle

%
\vspace{-5.0mm}  
\begin{abstract}
We describe a novel approach for computing collision-free \emph{global} trajectories 
for $p$ agents with specified initial and final
configurations, based on an improved version of the alternating 
direction method of multipliers (ADMM). 
Compared with existing methods, our approach is naturally parallelizable and allows for 
incorporating
different cost functionals with only minor adjustments.
We apply our method to classical challenging instances
and observe that its computational requirements scale
well with $p$ for several cost functionals.
We also show that a specialization of our algorithm can be used for {\em local} 
motion planning 
by solving the problem of joint optimization in velocity space.
\end{abstract}

\vspace{-3mm} 
\section{Introduction}


Robot navigation relies on at least three sub-tasks: localization, mapping, and motion planning.  
The latter can be described as an optimization problem: compute the lowest-cost path, or \emph{trajectory}, between an initial and final configuration. 
This paper focuses on trajectory planning for {\em multiple} agents, an important problem in robotics~\cite{alonsomora12ijrr, wurman08}, computer animation, and crowd simulation~\cite{guy09}.

Centralized planning for multiple agents is PSPACE hard~\cite{reif79,hopcroft84}. 
To contend with this complexity, traditional multi-agent planning prioritizes agents and computes their trajectories sequentially~\cite{bennewitz02}, leading to suboptimal solutions.
By contrast, our method plans for all agents simultaneously.
Trajectory planning is also simplified if agents are non-distinct and can be dynamically assigned to a set of goal positions~\cite{alonsomora12ijrr}. 
We consider the harder problem where robots have a unique identity and their goal positions are statically pre-specified.
Both mixed-integer quadratic programming (MIQP)~\cite{mellinger12} and [more efficient, although local] sequential convex programming~\cite{augugliaro12} approaches have been applied to the problem of computing collision-free trajectories for multiple agents with pre-specified goal positions; however, due to the non-convexity of the problem, these approaches, especially the former, do not scale well with the number of agents. 
Alternatively, trajectories may be found by sampling in their joint configuration space~\cite{lavalle01}. This approach is probabilistic and, alone, only gives asymptotic guarantees.
See Appendix \ref{app:more_on_literature} for further comments on discrete search methods.

Due to the complexity of planning collision-free trajectories, \emph{real-time} robot navigation is commonly decoupled into a global planner and a fast local planner that performs collision-avoidance. 
Many single-agent reactive collision-avoidance algorithms are based either on potential fields~\cite{khatib86}, which typically ignore the velocity of other agents, or ``velocity obstacles''~\cite{fiorini98}, which provide improved performance in dynamic environments by formulating the optimization in velocity space instead of Cartesian space.
Building on an extension of the velocity-obstacles
approach, recent work on centralized collision avoidance \cite{alonsomora13icra} computes collision-free local motions for all agents whilst maximizing a joint utility using either a computationally expensive MIQP or an efficient, though local, QP. 
While not the main focus of this paper, we show that a specialization of our approach to global-trajectory optimization also applies for local-trajectory optimization, and our numerical results demonstrate improvements in both efficiency and scaling performance.


In this paper we formalize the global trajectory planning task as follows.
Given $p$ agents of different radii $\{r_i\}^p_{i = 1}$ with given desired initial and final positions, $\{x_i(0)\}^p_{i = 1}$ and $\{x_i(T)\}^p_{i = 1}$, along with a cost functional over trajectories, compute collision-free trajectories for all agents that minimize the cost functional. 
That is, find a set of intermediate points $\{x_i(t)\}^p_{i = 1}$, $t \in (0, T)$, that satisfies the ``hard'' collision-free constraints that $\|x_i(t) - x_j(t) \| > r_i + r_j$, for all $i$, $j$ and $t$, and that insofar as possible, minimizes the cost functional.

The method we propose searches for a solution within the space of piece-wise linear trajectories, wherein the trajectory of an agent is completely specified by a set of positions at a fixed set of time instants $\{t_s \}^\n_{s = 0}$. 
We call these time instants \emph{break-points} and they are the same for all agents, which greatly simplifies the mathematics of our method. 
All other intermediate points of the trajectories are computed by assuming that each agent moves with constant velocity in between break-points: if $t_1$ and $t_2 > t_1$ are consecutive break-points, then $x_i(t) = \frac{1}{t_2 - t_1}((t_2-t)x_i(t_1) + (t-t_1)x_i(t_2))$ for $t \in [t_1, t_2]$.
Along with the set of initial and final configurations, the number of interior break-points ($\n-1$) is an input to our method, with a corresponding tradeoff: increasing $\n$ yields trajectories that are more flexible and smooth, with possibly higher quality; but increasing $\n$ enlarges the problem, leading to potentially increased computation.

The main contributions of this paper are as follows:
\begin{enumerate}[i)]
\item We formulate the global trajectory planning task as a decomposable optimization problem. We show how to solve the resulting sub-problems exactly and efficiently, despite their non-convexity, and how to coordinate their solutions using message-passing. Our method, based on the ``three-weight'' version of ADMM \cite{nate13}, is easily parallelized, does not require parameter tuning, and we present empirical evidence of good scalability with $p$.
\item Within our decomposable framework, we describe different sub-problems, called {\em minimizers}, each ensuring the trajectories satisfy a separate criterion.
Our method is flexible and can consider different combinations of minimizers. 
A particularly crucial minimizer ensures there are no inter-agent collisions, but we also  derive other minimizers that allow for finding trajectories with minimal total energy, avoiding static obstacles, or imposing dynamic constraints, such as maximum/minimum agent velocity.
\item We show that our method can specialize to perform local planning by solving the problem of joint optimization in velocity space~\cite{alonsomora13icra}.
\end{enumerate}

Our work is among the few examples where the success of applying ADMM to find approximate solutions to a large non-convex problems can be judged with the naked eye, by the gracefulness of the trajectories found. This paper also reinforces the claim in \cite{nate13} that small, yet important, modifications to ADMM can bring an order of magnitude increase in speed. We emphasize the importance of these modifications in our numerical experiments, where we compare the performance of our method using the three-weight algorithm (TWA) versus that of standard ADMM.

The rest of the paper is organized as follows.
Section \ref{sec:back} provides background on ADMM and the TWA. 
Section \ref{sec:global} formulates the global-trajectory-planning task as an optimization problem and describes the separate blocks necessary to solve it (the mathematical details of solving these sub-problems are left to appendices).
Section \ref{sec:num} evaluates the performance of our solution: its scalability with $p$, sensitivity to initial conditions, and the effect of different cost functionals. 
Section \ref{sec:co} explains how to implement a velocity-obstacle method using our method and compares its performance with prior work.
Finally, Section \ref{sec:future} draws conclusions and suggests directions for future work.

\section{Minimizers in the TWA} \label{sec:back}

In this section we provide a short description of the TWA \cite{nate13}, and, in particular, the role of the {\em minimizer} building blocks that it needs to solve a particular optimization problem.
Section \ref{app:mod_mess_pass_admm} of the supplementary material includes a full description of the TWA.

As a small illustrative example of how the TWA is used to solve optimization problems, suppose we want to solve $\min_{x \in \reals^3} f(x) = \min_{\{x_1,x_2,x_3\}} f_1(x_1,x_3) + f_2(x_1,x_2,x_3) + f_3(x_3)$, where $f_i(.) \in \reals \cup \{+\infty\}$. 
The functions can represent soft costs, for example $f_3(x_3) = (x_3 - 1)^2$, or hard equality or inequality constraints, such as  $f_1(x_1, x_3) = \mathbb{J}(x_1\leq x_3)$, where we are using the notation $\mathbb{J}(.) = 0$ if $(.)$ is true or $+\infty$ if $(.)$ is false. 

\begin{figure}[h!]
\begin{center}
\includegraphics[width=0.45\textwidth]{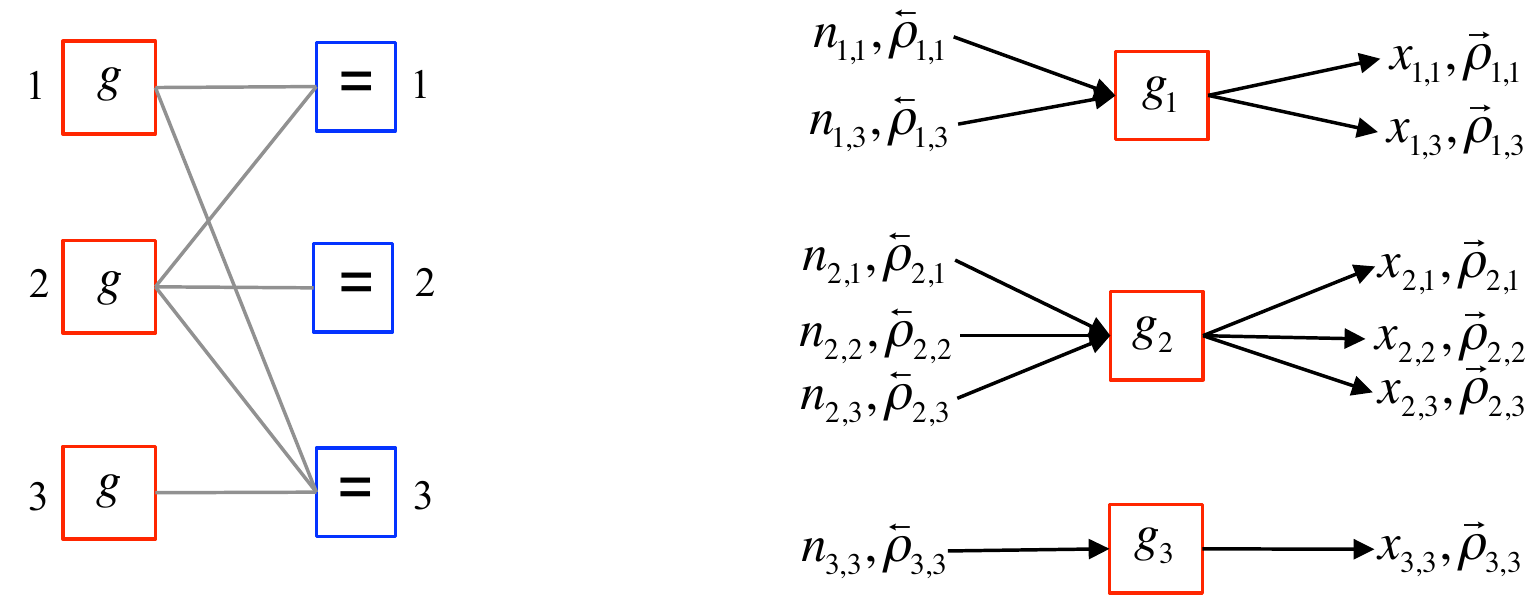}
\caption{Left: bipartite graph, with one minimizer-node on the left for each function making up the overall objective function, and one equality-node on the right per variable in the problem. Right: The input and output variables for each minimizer block.}
\label{fig:bipartite_simple_example}
\end{center}
\end{figure}

The TWA solves this optimization problem iteratively by passing messages on a bipartite graph, in the form of a Forney factor graph \cite{Forney}: one \emph{minimizer-node} per function $f_b$, one \emph{equality-node} per variable $x_j$ and an edge $(b,j)$, connecting $b$ and $j$, if $f_b$ depends on $x_j$ (see Figure \ref{fig:bipartite_simple_example}-left).

Apart from the first-iteration message values, and two internal parameters\footnote{These are the step-size and $\rho_0$ constants. See Section \ref{app:mod_mess_pass_admm} in the supplementary material for more detail.} that we specify in Section \ref{sec:num}, the algorithm is fully specified by the behavior of the minimizers and the topology of the graph.

What does a minimizer do? The minimizer-node $g_1$, for example, solves a small optimization problem over its local variables $x_1$ and $x_3$. 
Without going into the full detail presented in \cite{nate13} and the supplementary material, the estimates $x_{1,1}$ and $x_{1,3}$ are then combined with running sums of the differences between the minimizer estimates and the equality-node consensus estimates to obtain {\em messages} $m_{1,1}$ and $m_{1,3}$ on each neighboring edge that are sent to the neighboring equality-nodes along with corresponding \emph{certainty weights}, $\overrightarrow{\rho}_{1,2}$ and $\overrightarrow{\rho}_{1,3}$. 
All other minimizers act similarly. 

The equality-nodes receive these local messages and weights and produce consensus estimates for all variables by computing an average of the incoming messages, weighted by the incoming certainty weights $\overrightarrow{\rho}$. 
From these consensus estimates, {\em correcting messages} are computed and communicated back to the minimizers to help them reach consensus. 
A certainty weight for the correcting messages, $\overleftarrow{\rho}$, is also communicated back to the minimizers. 
For example, the minimizer $g_1$ receives correcting messages $n_{1,1}$ and $n_{1,3}$ with corresponding certainty weights $\overleftarrow{\rho}_{1,1}$ and $\overleftarrow{\rho}_{1,3}$ (see Figure \ref{fig:bipartite_simple_example}-right).

When producing new local estimates, the $b$th minimizer node computes its local estimates $\{x_j\}$ by choosing a point that minimizes the sum of the local function $f_b$ and weighted squared distance from the incoming messages (ties are broken randomly):
\begin{equation}\label{eq:general_minimizer_definition}
\{x_{b,j}\}_j = g_b \left(\{n_{b,j}\}_j,\{\overleftarrow{\rho}^{k}_{b,j}\}_j\right)
\equiv \arg \min_{ \{x_j\}_j  } \left[ f_b( \{x_j\}_j   ) +\frac{1}{2} \sum_j \overleftarrow{\rho}_{b,j} (x_j - n_{b,j})^2 \right],
\end{equation}
where $\{ \}_j$ and $\sum_j$ run over all equality-nodes connected to $b$. 
In the TWA, the certainty weights $\{ \overrightarrow{\rho}_{b,j}\}$ that this minimizer outputs must be $0$ (\emph{uncertain}); $\infty$ (\emph{certain}); or $\rho_0$, set to some fixed value. 
The logic for setting weights from minimizer-nodes depends on the problem; as we shall see, in trajectory planning problems, we only use $0$ or $\rho_0$ weights. 
If we choose that all minimizers always output weights equal to $\rho_0$, the TWA reduces to standard ADMM; however, 0-weights allows equality-nodes to ignore inactive constraints, traversing the search space much faster.

Finally, notice that all minimizers can operate simultaneously, and the same is true for the consensus calculation performed by each equality-node. 
The algorithm is thus easy to parallelize. 

\section{Global trajectory planning} \label{sec:global}

We now turn to describing our decomposition of the global trajectory planning optimization problem in detail.
We begin by defining the variables to be optimized in our optimization problem.
In our formulation, we are \emph{not} tracking the points of the trajectories by a continuous-time variable taking values in $[0,T]$.
Rather, our variables are the positions $\{x_i(s)\}_{i \in [p]}$,
where the trajectories are indexed by $i$ and break-points are 
indexed by a discrete variable $s$ taking values between $1$ and $\n-1$. 
Note that $\{x_i(0)\}_{i \in [p]}$ and $\{x_i(\n)\}_{i \in [p]}$ are the initial and final configuration, sets of fixed values, not variables to optimize. 
%
%

\subsection{Formulation as unconstrained optimization without static obstacles}

%
In terms of these variables, the {\em non-collision} constraints\footnote{We replaced the strict inequality in the condition for non-collision by a simple inequality ``$\geq$'' to avoid technicalities in formulating the optimization problem. Since the agents are round, this allows for a single point of contact between two agents and does not reduce practical relevance.} are 
\begin{align}\label{eq:basic_opt_prob}
& \| (\alpha x_i(s+1) + (1 - \alpha) x_i(s)) - (\alpha x_j(s+1) + (1 - \alpha) x_j(s)) \| \geq r_i + r_j,
\\
& \text{for all }\; i,j \in [p], s \in \{0,...,\n-1\} \text{ and } \alpha \in [0,1].\nonumber
\end{align}
The parameter $\alpha$ is used to trace out the constant-velocity trajectories of agents $i$ and $j$ between break-points $s+1$ and $s$. 
The parameter $\alpha$ has no units, it is a normalized time rather than an absolute time. 
If $t_1$ is the absolute time of the break-point with integer index $s$ and $t_2$ is the absolute time of the break-point with integer index $s+1$ and $t$ parametrizes the trajectories in absolute time then $\alpha = (t-t_1)/(t_2 - t_1)$. 
Note that in the above formulation, absolute time does not appear, and any solution is simply a set of paths that, when travelled by each agent at constant velocity between break-points, leads to no collisions. 
When converting this solution into trajectories parameterized by absolute time, the break-points do not need to be chosen uniformly spaced in absolute time.

The constraints represented in (\ref{eq:basic_opt_prob}) can be formally incorporated into an unconstrained optimization problem as follows. 
We search for a solution to the problem:
\begin{equation} \label{eq:unconst_opt}
\underset{\{x_i(s)\}_{i, s}}{\text{min}} f^{\text{cost}}(\{x_i(s)\}_{i, s}) + \sum^{\n-1}_{s =0} \sum_{i > j} f^{\text{coll}}_{r_i,r_j} (x_i(s),x_i(s+1),x_j(s),x_j(s+1)),
\end{equation}
where $\{x_i(0)\}_p$ and $\{x_i(\n)\}_p$ are constants rather than optimization variables, and where the function $f^{\text{cost}}$ is a function that represents some cost to be minimized (e.g. the integrated kinetic energy or the maximum velocity over all the agents) and the function $f^{\text{coll}}_{r,r'}$ is defined as,
\begin{equation} \label{eq:f_coll_def}
f^{\text{coll}}_{r,r'}(\ux,\ox,\ux',\ox') = \mathbb{J}( \| \alpha (\ox-\ox') + (1 - \alpha) (\ux - \ux') \| \geq r + r' \;\forall \alpha \in [0,1]).
\end{equation}
In this section, $\ux$ and $\ox$ represent the position of an arbitrary agent of radius $r$ at two consecutive break-points and $\ux'$ and $\ox'$ the position of a second arbitrary agent of radius $r'$ at the same break-points. 
In the expression above $\mathbb{J}(.)$ takes the value $0$ whenever its argument, a clause, is true and takes the value $+\infty$ otherwise.
Intuitively, we pay an infinite cost in $f^{\text{coll}}_{r,r'}$ whenever there is a collision, and we pay zero otherwise.

In \eqref{eq:unconst_opt} we can set $f^{\text{cost}}(.)$, to enforce a preference for trajectories satisfying specific properties. 
For example, we might prefer trajectories for which the total kinetic energy spent by the set of agents is small. 
In this case, defining $f^{\text{cost}}_C (\ux,\ox) = C \| \ux - \ox\|^2$, we have,
\begin{equation} \label{eq:velo_cost_pair_wise}
f^{\text{cost}}(\{x_i(s)\}_{i, s}) = \frac{1}{p \n} \sum^p_{i=1} \sum^{\n-1}_{s = 0} f^{\text{cost}}_{C_{i,s}}(x_i(s),x_i(s+1)).
\end{equation}
where the coefficients $\{C_{i,s}\}$ can account for agents with different masses, different absolute-time intervals between-break points or different preferences regarding which agents we want to be less active and which agents are allowed to move faster.

More simply, we might want to exclude trajectories in which agents move faster than a certain amount, but without distinguishing among all remaining trajectories. 
For this case we can write,
\begin{equation}\label{eq:max_vel_min}
f^{\text{cost}}_C (\ux,\ox) = \mathbb{J}(\|\ox - \ux\| \leq C).
\end{equation}
In this case, associating each break-point to a time instant, the coefficients $\{C_{i,s}\}$ in expression \eqref{eq:velo_cost_pair_wise} would represent different limits on the velocity of different agents between different sections of the trajectory. 
If we want to force all agents to have a minimum velocity we can simply reverse the inequality in \eqref{eq:max_vel_min}.

%
%

\subsection{Formulation as unconstrained optimization with static obstacles}

In many scenarios agents should also avoid collisions with static obstacles. 
Given two points in space, $x_L$ and $x_R$, we can forbid all agents from crossing the line segment from $x_L$ to $x_R$ by adding the following term to the function 
\eqref{eq:unconst_opt}: $\sum^p_{i=1} \sum^{\n-1}_{s=0} f^{\text{wall}}_{x_L,x_R,r_i}(x_i(s),x_i(s+1))$.
%
%
We recall that $r_i$ is the radius of agent $i$ and 
\begin{equation} \label{eq:wall_minimizer}
f^{\text{wall}}_{x_L,x_R,r}(\ux,\ox) = \mathbb{J}( \| (\alpha \ox + (1 - \alpha) \ux ) - (\beta x_R + (1-\beta) x_L)\| \geq r \text{ for all } \alpha,\beta \in [0,1]).
\end{equation}
Notice that $f^{\text{coll}}$ can be expressed using $f^{\text{wall}}$. 
In particular,
\begin{equation}\label{eq:wall_min_goes_to_coll_min}
f^{\text{coll}}_{r,r'}(\ux,\ox,\ux',\ox') = f^{\text{wall}}_{0,0,r+r'}(\ux'-\ux,\ox'-\ox).
\end{equation}
We use this fact later to express the minimizer associated with agent-agent collisions using the minimizer associated with agent-obstacle collisions.

When agents move in the plane, i.e. $x_i(s) \in \reals^2$ for all $i \in [p]$ and $s+1 \in  [\n+1]$, being able to avoid collisions with a general static line segment allows to automatically avoid collisions with multiple static obstacles of arbitrary polygonal shape. 
Our numerical experiments only consider agents in the plane and so, in this paper, we only describe the minimizer block for wall collision for a 2D world. 
In higher dimensions, different obstacle primitives need to be considered. 

%
%

\subsection{Message-passing formulation}\label{sec:mess_pass}

To solve \eqref{eq:unconst_opt} using the TWA, we need to specify the topology of the bipartite graph associated with the unconstrained formulation \eqref{eq:unconst_opt} and the operation performed by every
minimizer, i.e. the $\overrightarrow{\rho}$-weight update logic and $x$-variable update equations. 
We postpone describing the choice of initial values and internal parameters until Section \ref{sec:num}.

We first describe the bipartite graph.
To be concrete, let us assume that the cost functional has the form of \eqref{eq:velo_cost_pair_wise}.
The unconstrained formulation \eqref{eq:unconst_opt} then tells us that the global objective function is the sum of $\n p ( p + 1) /2$ terms: $\n p (p-1)/2$ functions $f^{\text{coll}}$ and $\n p$ functions $f^{\text{cost}}_C$.
These functions involve a total of $(\n+1) p$ variables out of which only $(\n-1) p$ are free (since the initial and final configurations are fixed). 
Correspondingly, the bipartite graph along which messages are passed has $\n p ( p + 1) /2$ minimizer-nodes that connect to the $(\n+1)p$ equality-nodes. 
In particular, the equality-node associated with the break-point variable $x_i(s)$, $\n>s>0$, is connected to $2(p-1)$ different $g^{\text{coll}}$ minimizer-nodes and two different $g^{\text{cost}}_C$ minimizer-nodes. 
If $s = 0$ or $s=\n$ the equality-node only connects to half as many $g^{\text{coll}}$ nodes and $g^{\text{cost}}_C$ nodes.

We now describe the different minimizers.
Every minimizer basically is a special case of \eqref{eq:general_minimizer_definition}.

%
%

\subsubsection{Agent-agent collision minimizer}

We start with the minimizer associated with the functions $f^{\text{coll}}$, that we denoted by $g^{\text{coll}}$. 
This minimizer receives as parameters the radius, $r$ and $r'$, of the two agents whose collision it is avoiding. 
The minimizer takes as input a set of incoming $n$-messages, $\{\un,\on,\un',\on'\}$, and associated $\overleftarrow{\rho}$-weights, $\{\urho,\orho,\urho',\orho'\}$, and outputs a set of updated $x$-variables according to expression \eqref{eq:coll_minimizer_problem}.  
Messages $\un$ and $\on$ come from the two equality-nodes associated with the positions of one of the agents at two consecutive break-points and $\un'$ and $\on'$ from the corresponding equality-nodes for the other agent.
\begin{align} \label{eq:coll_minimizer_problem}
&g^{\text{coll}}(\un,\on,\un',\on',\urho,\orho,\urho',\orho',r,r') =  \arg \min_{\{\ux,\ox,\ux',\ox'\}} f^{\text{coll}}_{r,r'}(\ux,\ox,\ux',\ox')\nonumber\\
& + \frac{\urho}{2} \|\ux - \un \|^2 + \frac{\orho}{2} \|\ox - \on \|^2 + \frac{\urho'}{2} \|\ux' - \un' \|^2 + \frac{\orho'}{2} \|\ox' - \on' \|^2.
\end{align}
The update logic for the weights $\overrightarrow{\rho}$ for this minimizer is simple. 
If the trajectory from $\un$ to $\on$ for an agent of radius $r$ does not collide with the trajectory from $\un'$ to $\on'$ for an agent of radius $r'$ then set all the outgoing weights $\overrightarrow{\rho}$ to zero. 
Otherwise set them all to $\rho_0$. 
The outgoing zero weights indicate to the receiving equality-nodes in the bipartite graph that the collision constraint for this pair of agents is inactive and that the values it receives from this minimizer-node should be ignored when computing the consensus values of the receiving equality-nodes.

The solution to \eqref{eq:coll_minimizer_problem} is found using the agent-obstacle collision minimizer that we describe next.

%
%

\subsubsection{Agent-obstacle collision minimizer}

The minimizer for $f^{\text{wall}}$ is denoted by $g^{\text{wall}}$. 
It is parameterized by the obstacle position $\{x_L,x_R\}$ as well as the radius of the agent that needs to avoid the obstacle. 
It receives two $n$-messages, $\{\un,\on\}$, and corresponding weights $\{\urho,\orho\}$, from the equality-nodes associated with two consecutive positions of an agent that needs to avoid the obstacle. 
Its output, the $x$-variables, are defined as
\vspace{-1em} 
\begin{equation} \label{eq:wall_minimizer_problem}
g^{\text{wall}}(\un,\on,r,x_L,x_R,\urho,\orho) = \arg \min_{\{\ux,\ox\}} f^{\text{wall}}_{x_L,x_R,r}(\ux,\ox) + \frac{\urho}{2}\|\ux - \un \|^2 + \frac{\orho}{2}\|\ox -\on\|^2.
\end{equation}
When agents move in the plane (2D), this minimizer can be solved by reformulating the optimization in \eqref{eq:wall_minimizer_problem} as a mechanical problem involving a system of springs that we can solve exactly and efficiently.
This reduction is explained in the supplementary material in Section \ref{app:agent_obstacle_minimizer} and the solution to the mechanical problem is explained in Section \ref{app:mechanical_analog}.

The update logic for the $\overrightarrow{\rho}$-weights is similar to that of the $g^{\text{coll}}$ minimizer.
If an agent of radius $r$ going from $\un$ and $\on$ does not collide with the line segment from $x_L$ to $x_R$ then set all outgoing weights to zero because the constraint is inactive; otherwise set all the outgoing weights to $\rho_0$.

Notice that, from \eqref{eq:wall_min_goes_to_coll_min}, it follows that the agent-agent minimizer $g^{\text{coll}}$ can be expressed using $g^{\text{wall}}$. More concretely, as proved in the supplementary material, Section \ref{app:agent_agent_minimizer},
\begin{align*}
g^{\text{coll}}(\un,\on,\un',\on',\urho,\orho,\urho',\orho',r,r')= M_2 g^{\text{wall}}\left(M_1 . \{\un,\on,\un',\on',\urho,\orho,\urho',\orho',r,r'\} \right),
\end{align*}
for a constant rectangular matrix $M_1$ and a matrix $M_2$ that depend on $\{\un,\on,\un',\on',\urho,\orho,\urho',\orho'\}$.

%
%

\subsubsection{Minimum energy and maximum (minimum) velocity minimizer}

When $f^{\text{cost}}$ can be decomposed as in \eqref{eq:velo_cost_pair_wise}, the minimizer associated with the functions $f^{\text{cost}}$ is denoted by $g^{\text{cost}}$ and receives as input two $n$-messages, $\{\un,\on\}$, and corresponding weights, $\{\urho,\orho\}$. 
The messages come from two equality-nodes associated with two consecutive positions of an agent. 
The minimizer is also parameterized by a cost factor $c$. 
It outputs a set of updated $x$-messages defined as
\begin{align}
g^{\text{cost}}(\un,\on,\urho,\orho,c) =  \arg \min_{\{\ux,\ox\}} f^{\text{cost}}_{c}(\ux,\ox)+ \frac{\urho}{2} \|\ux - \un \|^2 + \frac{\orho}{2} \|\ox - \on \|^2.
\end{align}
%
The update logic for the $\overrightarrow{\rho}$-weights of the minimum energy minimizer is very simply: always set all outgoing weights $\overrightarrow{\rho}$ to $\rho_0$.
The update logic for the $\overrightarrow{\rho}$-weights of the maximum velocity minimizer is the following. If $\| \un  - \on\| \leq c$ set all outgoing weights to zero. Otherwise, set them to $\rho_0$.
The update logic for the minimum velocity minimizer is similar.
If $\| \un  - \on\| \geq c$, set all the $\overrightarrow{\rho}$-weights to zero. Otherwise set them to $\rho_0$.

The solution to the minimum energy, maximum velocity and minimum velocity minimizer is written in the supplementary material in Sections \ref{app:energy_minimizer}, \ref{app:max_vel_minimizer}, and \ref{app:min_vel_minimizer} respectively. 

%
%

\section{Numerical results} \label{sec:num}

We now report on the performance of our algorithm (see Appendix \ref{app:more_on_the_algorithm} for an important comment on the anytime
properties of our algorithm).
Note that the lack of open-source scalable algorithms for global trajectory planning in the literature makes it difficult to benchmark our performance against other methods.
Also, in a paper it is difficult to appreciate the gracefulness of the discovered trajectory optimizations, so we include a video in the supplementary material that shows final optimized trajectories as well as intermediate results as the algorithm progresses for a variety of additional scenarios, including those with obstacles.
All the tests described here are for agents in a two-dimensional plane.
All tests but the last were performed using six cores of a 3.4GHz i7 CPU.

The different tests did not require any special tuning of parameters. 
In particular, the step-size in \cite{nate13} (their $\alpha$ variable) is always $0.1$. 
In order to quickly equilibrate the system to a reasonable set of variables and to wash out the importance of initial conditions, the default weight $\rho_0$ was set equal to a small value ($\n p \times 10^{-5} $) for the first 20 iterations and then set to $1$ for all further iterations.

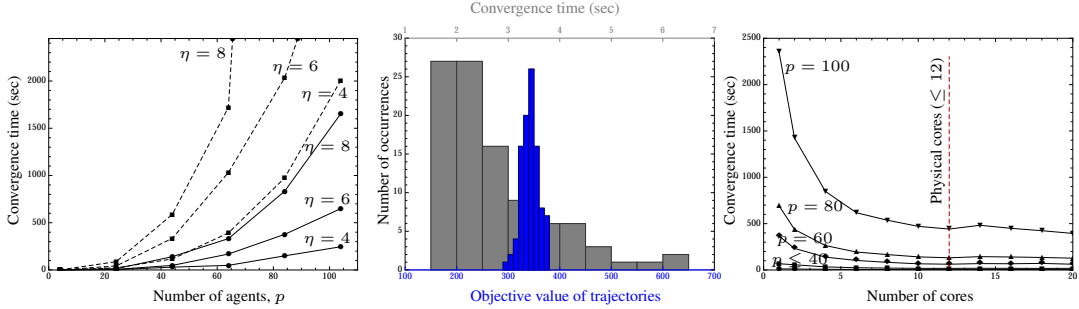
\begin{figure}[h!]
\begin{center}
\includegraphics[width=0.315\textwidth]{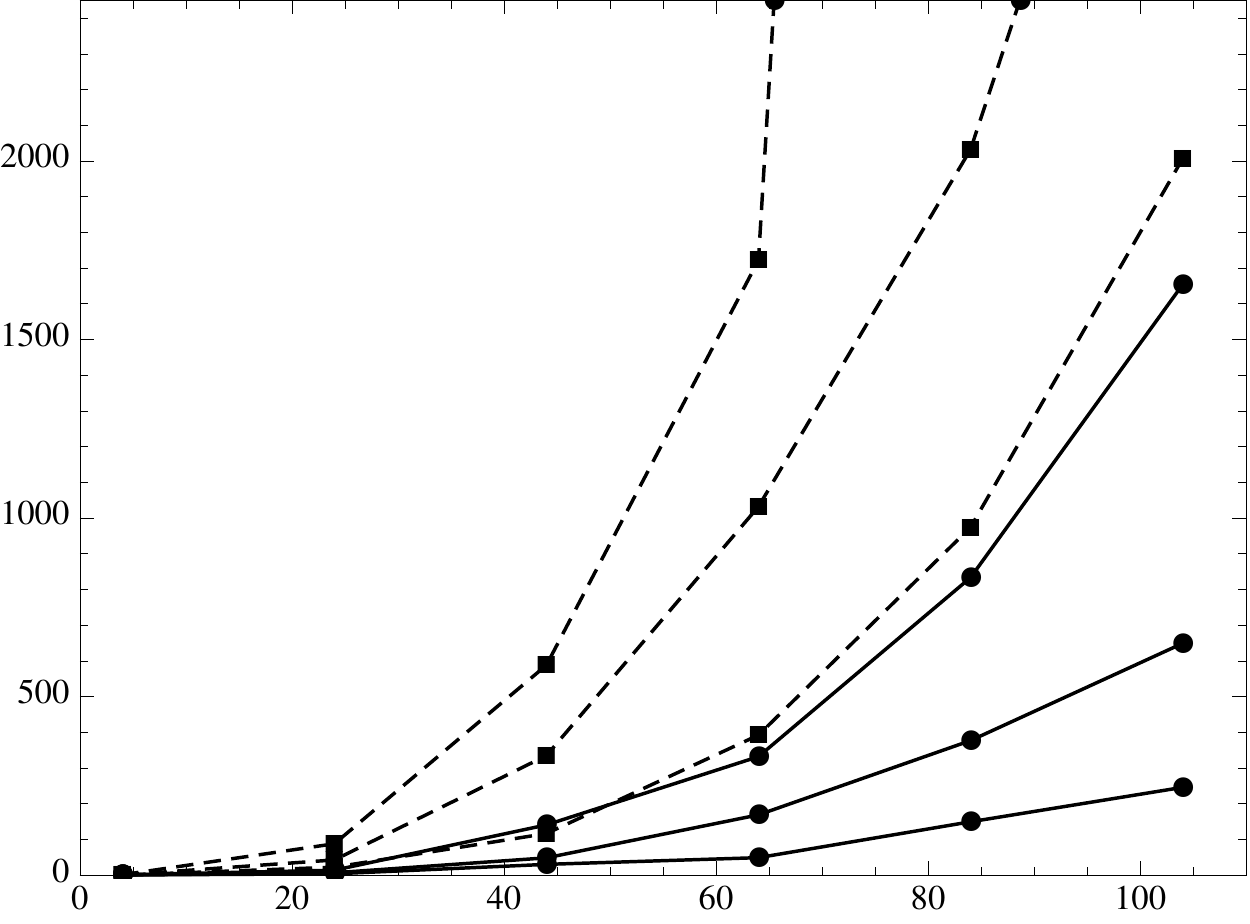}
\;\;\;
\includegraphics[width=0.313\textwidth]{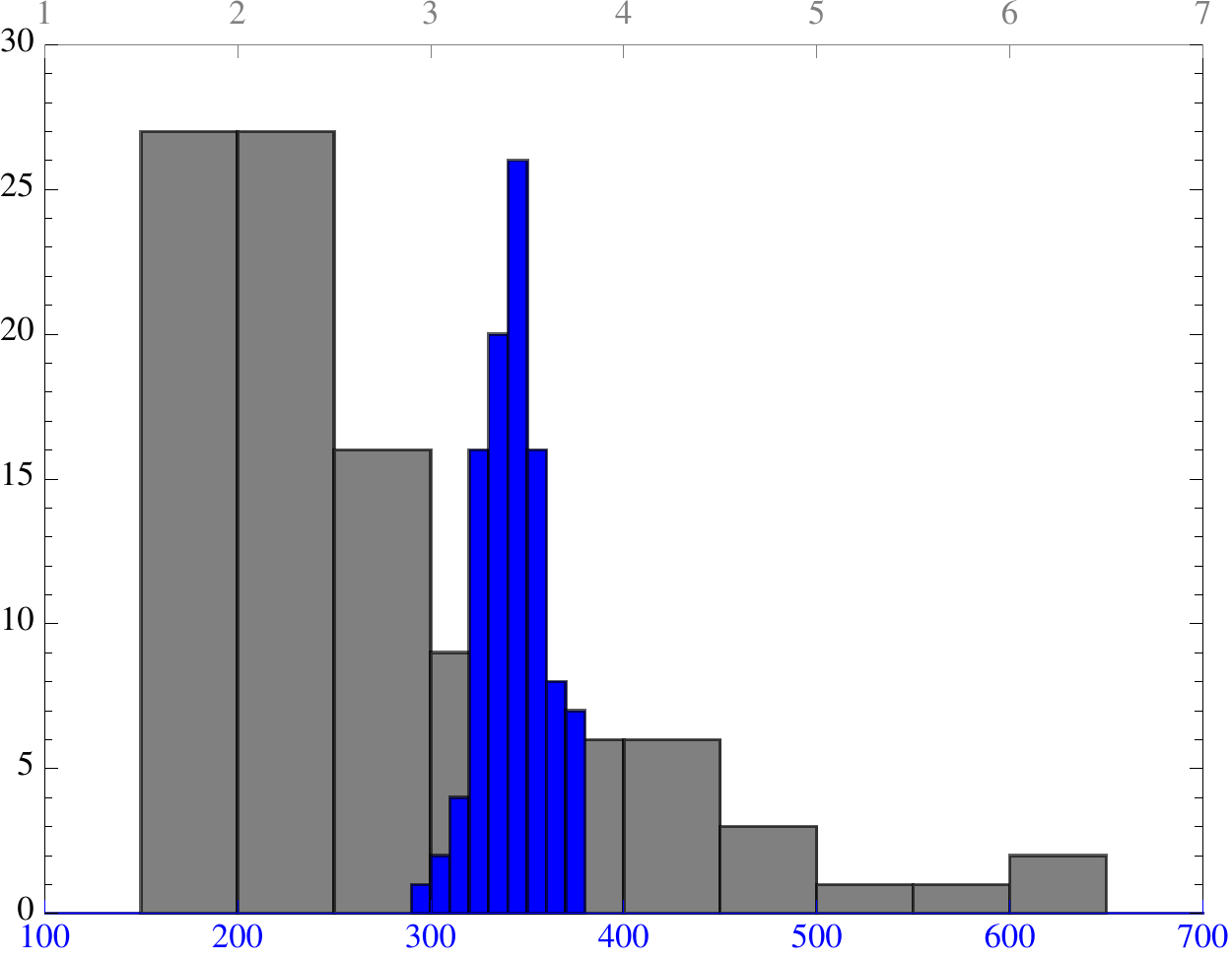}
\;
\includegraphics[width=0.32\textwidth]{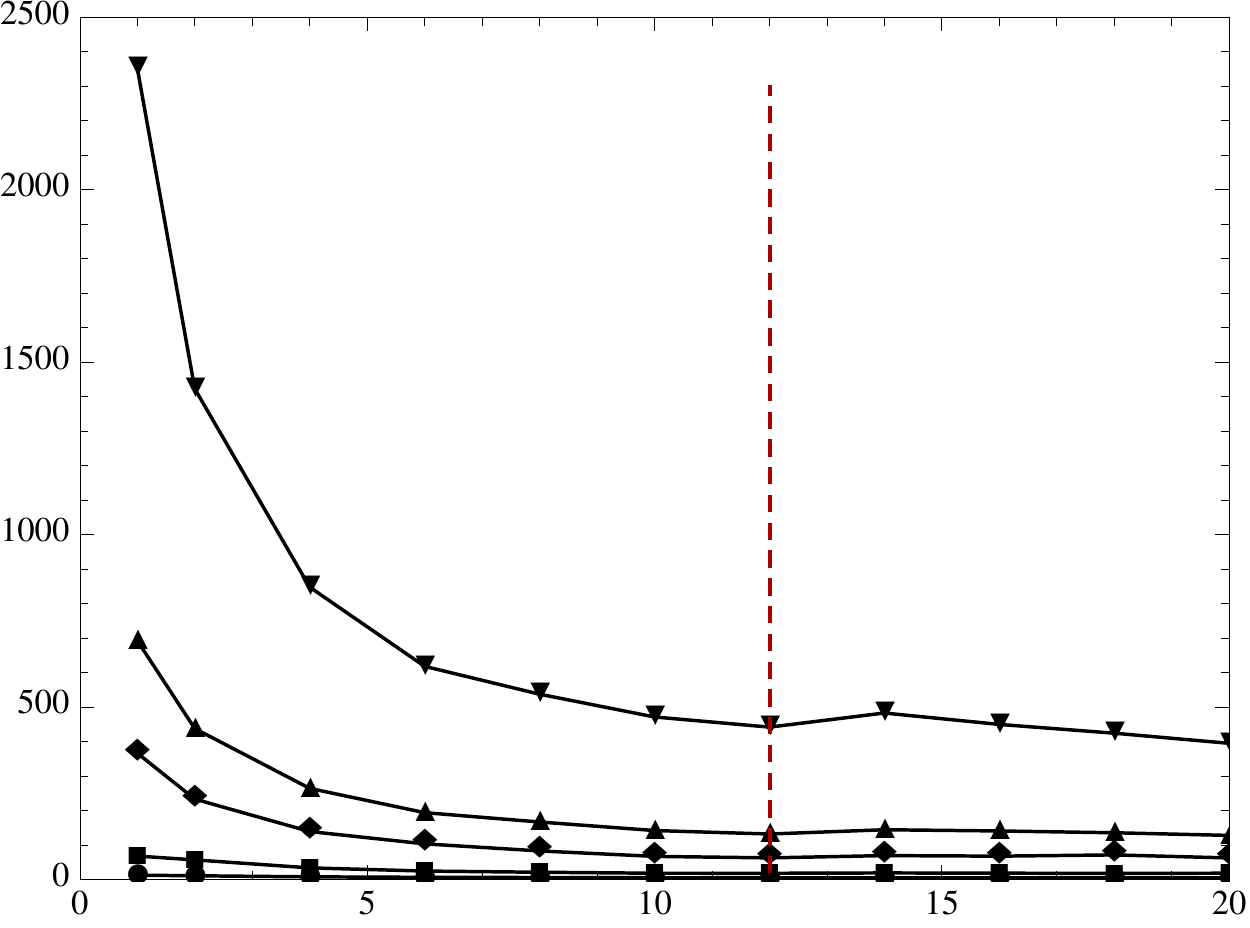}
\put(-134,22){\rotatebox{90}{\tiny Convergence time (sec)}}
\put(-80,-7){\rotatebox{0}{\tiny Number of cores}}
\put(-265,22){\rotatebox{90}{\tiny Number of occurrences}}
\put(-230,-7){\rotatebox{0}{\tiny \color{blue} Objective value of trajectories}}
\put(-230,102){\rotatebox{0}{\tiny \color{gray} Convergence time (sec)}}
\put(-350,-7){\rotatebox{0}{\tiny Number of agents, $p$}}
\put(-405,22){\rotatebox{90}{\tiny Convergence time (sec)}}
\put(-340,85){\rotatebox{0}{\tiny $\n = 8 $}}
\put(-305,80){\rotatebox{0}{\tiny $\n = 6 $}}
\put(-294,70){\rotatebox{0}{\tiny $\n = 4 $}}
\put(-293,50){\rotatebox{0}{\tiny $\n = 8 $}}
\put(-293,30){\rotatebox{0}{\tiny $\n = 6 $}}
\put(-293,15){\rotatebox{0}{\tiny $\n = 4 $}}
\put(-56,30){\rotatebox{90}{\tiny Physical cores ($\leq 12$)}}
\put(-111,80){\rotatebox{00}{\tiny $p = 100$}}
\put(-110,27){\rotatebox{00}{\tiny $p = 80$}}
\put(-114,15){\rotatebox{00}{\tiny $p = 60$}}
\put(-116,7){\rotatebox{00}{\tiny $p \leq 40$}}
\caption{Left: Convergence time using standard ADMM (dashed lines) and using TWA (solid lines). Middle: Distribution of total energy and time for convergence with random initial conditions ($p=20$ and $\n = 5$). Right: Convergence time using a different number of cores ($\n = 5$).}
\label{fig:admm_vs_dynamic_admm_histogram_cores}
\end{center}
\end{figure}

The first test considers scenario CONF1: $p$ (even) agents of radius $r$, equally spaced around on a circle of radius $R$, are each required to exchange position with the corresponding antipodal agent, $r = (5/4) R \sin(\pi/ 2(p -4))$. 
This is a classical difficult test scenario because the straight line motion of all agents to their goal would result in them all colliding in the center of the circle. 
We compare the convergence time of the TWA with a similar version using standard ADMM to perform the optimizations. 
In this test, the algorithm's initial value for each variable in the problem was set to the corresponding initial position of each agent. 
The objective is to minimize the total kinetic energy  ($C$ in the energy minimizer is set to $1$). 
Figure \ref{fig:admm_vs_dynamic_admm_histogram_cores}-left shows that the TWA scales better with $p$ than classic ADMM and typically gives an order of magnitude speed-up.
Please see Appendix \ref{app:more_on_the_convergence_time} for a further comment on the scaling of the convergence time of ADMM and TWA with $p$.
%
\vspace{-2.5mm} 

\vspace{-1.5em}
The second test for CONF1 analyzes the sensitivity of the convergence time and objective value when the variables' value at the first iteration are chosen uniformly at random in the smallest space-time box that includes the initial and final configuration of the robots. 
Figure \ref{fig:admm_vs_dynamic_admm_histogram_cores}-middle shows that, although there is some spread on the convergence time, our algorithm seems to reliably converge to relatively similar-cost local minima (other experiments show that the objective value of these minima is around 5 times smaller than that found when the algorithm is run using only the collision avoidance minimizers without a kinetic energy cost term). 
As would be expected, the precise trajectories found vary widely between different random runs.

Still for CONF1, and fixed initial conditions, we parallelize our method using several cores of a 2.66GHz i7 processor and a very primitive scheduling/synchronization scheme. 
Although this scheme does not fully exploit parallelization, Figure \ref{fig:admm_vs_dynamic_admm_histogram_cores}-right does show a speed-up as the number of cores increases and the larger $p$ is, the greater the speed-up. 
We stall when we reach the twelve physical cores available and start using virtual cores.

\begin{figure}[h!]
\begin{center}
\includegraphics[width=0.315\textwidth]{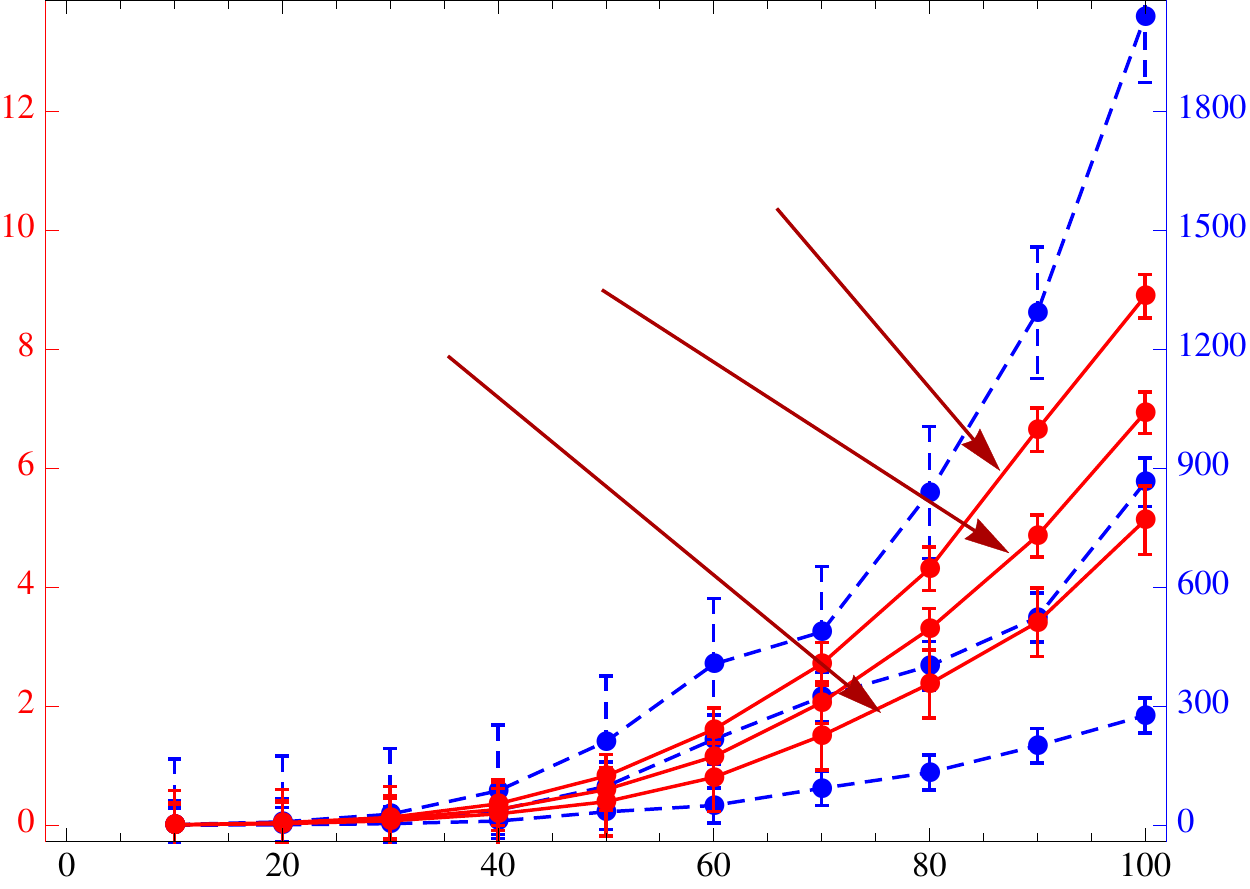}
\;\;\;\;\;\;\;\;\;\;\;\;\;\;\;\;\;\;\;\;\;\;\;
\includegraphics[trim = 0mm -1.0mm 00mm 0mm, clip,width=0.29\textwidth]{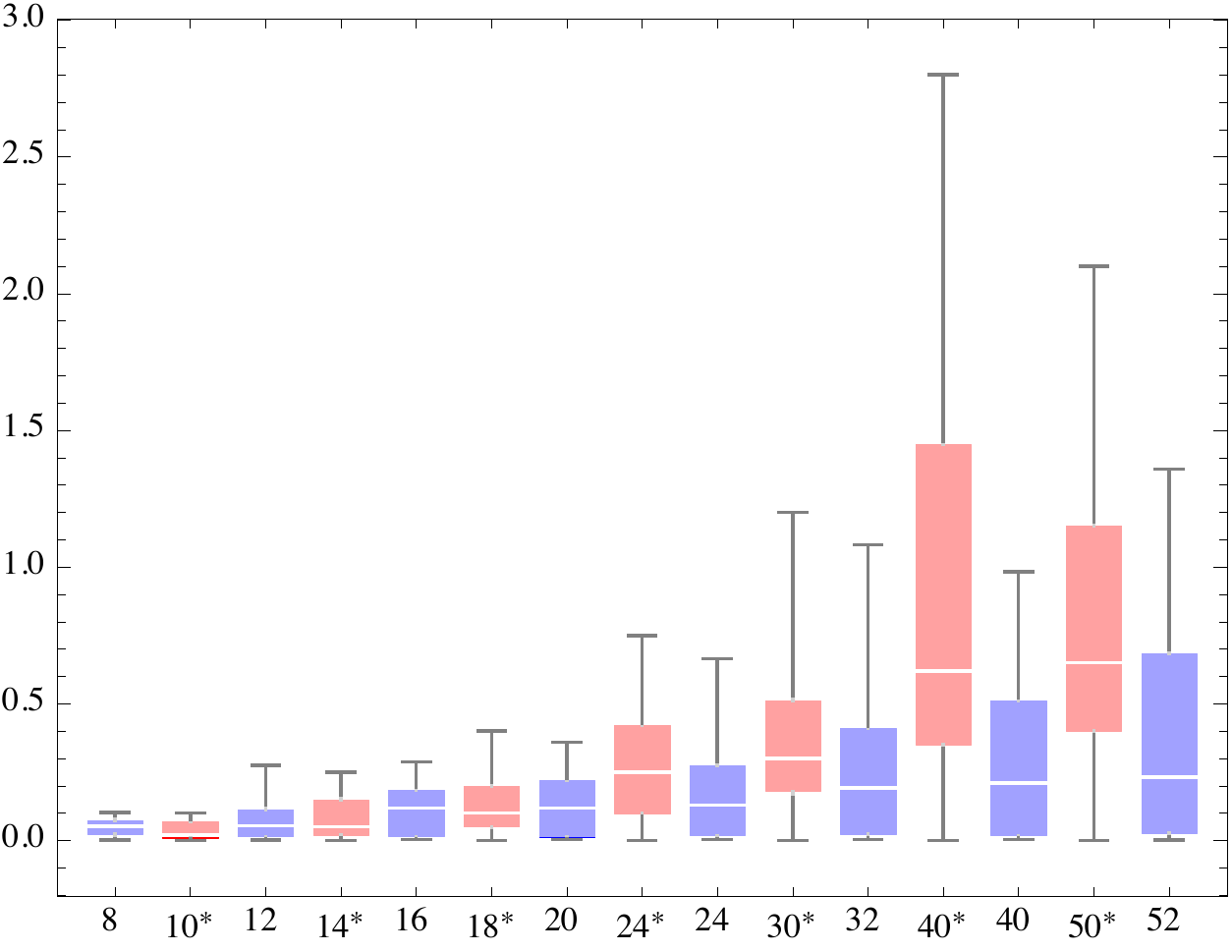}
\put(-328,29){\rotatebox{90}{\small \color{red} Feasible}}
\put(-318,15){\rotatebox{90}{\tiny \color{red} Convergence time (sec)}}
\put(-174,76){\rotatebox{-90}{\small \color{blue} Minimum energy}}
\put(-182,76){\rotatebox{-90}{\tiny \color{blue} Convergence time (sec)}}
\put(-275,-7){\rotatebox{0}{\tiny Number of agents, $p$}}
\put(-215,80){\rotatebox{00}{\tiny \color{blue} $\n = 8$}}
\put(-215,20){\rotatebox{00}{\tiny \color{blue} $\n = 6$}}
\put(-215,10){\rotatebox{00}{\tiny \color{blue} $\n = 4$}}
\put(-235,70){\rotatebox{00}{\tiny \color{red} $\n = 8$}}
\put(-255,62){\rotatebox{00}{\tiny \color{red} $\n = 6$}}
\put(-273,55){\rotatebox{00}{\tiny \color{red} $\n = 4$}}
\put(-75,60){\rotatebox{0}{\tiny \color{blue!20!white} Light blue: TWA}}
\put(-75,70){\rotatebox{0}{\tiny \color{pink} Pink: MIQP}}
\put(-75,-7){\tiny Number of agents, $p$}
\put(-123,08){\rotatebox{90}{\tiny Convergence time, one epoch (sec)}}
\caption{Left: Convergence time when minimizing energy (blue scale/dashed lines) and to simply find a feasible solution (red scale/solid lines). Right: (For Section \ref{sec:co}). Convergence-time distribution for each epoch using our method (blue bars) and using the MIQP of \cite{alonsomora13icra} (red bars and star-values).}
\label{fig:feasible_vs_opt_energy_and_VO_speed}
\end{center}
\end{figure}
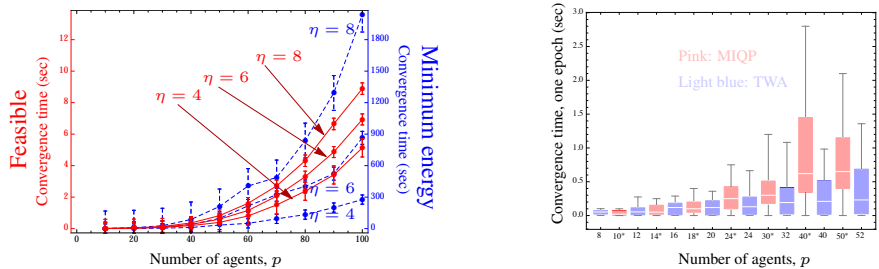

Finally, Figure \ref{fig:feasible_vs_opt_energy_and_VO_speed}-left compares the convergence time to optimize the total energy with the time to simply find a feasible (i.e. collision-free) solution. 
The agents initial and final configuration is randomly chosen in the plane (CONF2). Error bars indicate $\pm$ one standard deviation. Minimizing the kinetic energy is orders of magnitude computationally more expensive than finding a feasible solution, as is clear from the different magnitude of the left and right scale of Figure \ref{fig:feasible_vs_opt_energy_and_VO_speed}-left.
%
%
%
\section{Local trajectory planning based on velocity obstacles} \label{sec:co}

In this section we show how the joint optimization presented in~\cite{alonsomora13icra}, which is based on the concept of velocity obstacles~\cite{fiorini98} (VO), can be also solved via the message-passing TWA. 
In VO, given the current position $\{x_i(0)\}_{i \in [p]}$ and radius $\{r_i\}$ of all agents, a new velocity command is computed jointly for all agents minimizing the distance to their preferred velocity $\{v^{\text{ref}}_i\}_{i \in [p]}$. 
This new velocity command must guarantee that the trajectories of all agents remain collision-free for at least a time horizon $\tau$. 
New collision-free velocities are computed every $\alpha \tau$ seconds, $\alpha < 1$, until all agents reach their final configuration. 
Following~\cite{alonsomora13icra}, and assuming an obstacle-free environment and first order dynamics, the collision-free velocities are given by,
\begin{align*}
\underset{\{v_i\}_{i \in [p]}}{\text{minimize}} \sum_{i \in [p]} C_i \|v_i - v^{\text{ref}}_i \|^2 \text{ s.t. } \| (x_i(0) + v_i t) - (x_j(0) + v_j t) \| \geq r_i + r_j \;\; \forall \;\; i \in [p], t \in [0,\tau].
\end{align*}
Since the velocities $\{v_i\}_{i \in [p]}$ are related linearly to the final position of each object after $\tau$ seconds, $\{x_i(\tau)\}_{i \in [p]}$, a simple change of variables allows us to reformulate the above problem as,
\begin{align}\label{eq:rec_col_avoi_pos}
& \underset{\{x_i\}_{i \in [p]}}{\text{minimize}} \sum_{i \in [p]} C'_i \|x_i - x^{\text{ref}}_i \|^2 \nonumber\\
&\text{ s.t. } \| (1 - \alpha) (x_i(0) -x_j(0))+  \alpha (x_i - x_j)  \| \geq r_i + r_j \; \forall \; j> i \in [p], \alpha \in [0,1]
\end{align}
where $C'_i = C_i / \tau^2$, $x^{\text{ref}}_i = x_i(0) + v^{\text{ref}}_i \tau$ and we have dropped the $\tau$ in $x_i(\tau)$.
The above problem, extended to account for collisions with the static line segments $\{{x_R}_k,{x_L}_k\}_k$, can be formulated in an unconstrained form using the functions $f^{\text{cost}}$, $f^{\text{coll}}$ and $f^{\text{wall}}$. 
Namely, 
\begin{equation} \label{eq:unconst_opt_RVO}
\underset{\{x_i\}_{i}}{\text{min}} \sum_{i \in [p]} f^{\text{cost}}_{C'_i} (x_i,x^{\text{ref}}_i ) + \sum_{i > j} f^{\text{coll}}_{r_i,r_j} (x_i(0),x_i,x_j(0),x_j) + \sum_{i\in[p]} \sum_{k} f^{\text{wall}}_{{x_R}_k,{x_L}_k,r_i} (x_i(0),x_i).
\end{equation}
Note that $\{x_i(0)\}_i$ and $\{x^{\text{ref}}_i\}_i$ are constants, not variables being optimized.
Given this formulation, the TWA can be used to solve the optimization. 
All corresponding minimizers are special cases of minimizers derived in the previous section for global trajectory planning (see Section \ref{app:VO_minimizer} in the supplementary material for details).
Figure \ref{fig:feasible_vs_opt_energy_and_VO_speed}-right shows the distribution of the time to solve \eqref{eq:rec_col_avoi_pos} for CONF1.
We compare the mixed integer quadratic programming (MIQP) approach from~\cite{alonsomora13icra} with ours.
Our method finds a local minima of exactly  \eqref{eq:unconst_opt_RVO}, while \cite{alonsomora13icra} finds a global minima of an approximation to \eqref{eq:unconst_opt_RVO}. Specifically, \cite{alonsomora13icra} requires approximating the search domain by hyperplanes and an additional branch-and-bound algorithm while ours does not.
%
%
Both approaches use a mechanism for breaking the symmetry from CONF1 and avoid deadlocks: theirs uses a preferential rotation direction for agents, while we use agents with slightly different $C$ coefficients in their energy minimizers ($C_{i^{th} \text{ agent }} = 1 + 0.001 i$). 
Both simulations were done on a single 2.66GHz core. 
The results show the order of magnitude is similar, but, because our implementation is done in Java while \cite{alonsomora13icra} uses Matlab-mex interface of CPLEX 11, the results are not exactly comparable.

\section{Conclusion and future work} \label{sec:future}

We have presented a novel algorithm for global and local planning of the trajectory of multiple distinct agents, a problem known to be hard.
The solution is based on solving a non-convex optimization problem using TWA, a modified ADMM. Its similarity to ADMM brings scalability and easy parallelization. 
However, using TWA improves performance considerably. 
Our implementation of the algorithm in Java on a regular desktop computer, using a basic scheduler/synchronization over its few cores, already scales to hundreds of agents and achieves real-time performance for local planning.

%
The algorithm can flexibly account for obstacles and different cost functionals. For agents in the plane, we derived explicit expressions that account for static obstacles, moving obstacles, and dynamic constraints on the velocity and energy.  Future work should consider other restrictions on the smoothness of the trajectory (e.g. acceleration constraints) and provide fast solvers to our minimizers for agents in 3D. 

The message-passing nature of our algorithm hints that it might be possible to adapt our algorithm to do planning in a decentralized fashion. For example, minimizers like $g^{\text{coll}}$ could be solved by message exchange between pairs of agents within a maximum communication radius. It is an open problem to build a practical communication-synchronization scheme for such an approach.

%

\newpage
\renewcommand{\bibfont}{\footnotesize}
\bibliographystyle{unsrt}
\bibliography{ref}

\begin{thebibliography}{10}

\bibitem{alonsomora12ijrr}
Javier Alonso-Mora, Andreas Breitenmoser, Martin Rufli, Roland Siegwart, and
  Paul Beardsley.
\newblock Image and animation display with multiple mobile robots.
\newblock 31(6):753--773, 2012.

\bibitem{wurman08}
Peter~R. Wurman, Raffaello D'Andrea, and Mick Mountz.
\newblock Coordinating hundreds of cooperative, autonomous vehicles in
  warehouses.
\newblock {\em AI Magazine}, 29(1):9--19, 2008.

\bibitem{guy09}
Stephen~J. Guy, Jatin Chhugani, Changkyu Kim, Nadathur Satish, Ming Lin, Dinesh
  Manocha, and Pradeep Dubey.
\newblock Clearpath: highly parallel collision avoidance for multi-agent
  simulation.
\newblock In {\em Proceedings of the 2009 ACM SIGGRAPH/Eurographics Symposium
  on Computer Animation}, pages 177--187, 2009.

\bibitem{reif79}
John~H. Reif.
\newblock Complexity of the mover's problem and generalizations.
\newblock In {\em IEEE Annual Symposium on Foundations of Computer Science},
  pages 421--427, 1979.

\bibitem{hopcroft84}
John~E. Hopcroft, Jacob~T. Schwartz, and Micha Sharir.
\newblock On the complexity of motion planning for multiple independent
  objects; pspace-hardness of the "warehouseman's problem".
\newblock {\em The International Journal of Robotics Research}, 3(4):76--88,
  1984.

\bibitem{bennewitz02}
Maren Bennewitz, Wolfram Burgard, and Sebastian Thrun.
\newblock Finding and optimizing solvable priority schemes for decoupled path
  planning techniques for teams of mobile robots.
\newblock {\em Robotics and Autonomous Systems}, 41(2--3):89--99, 2002.

\bibitem{mellinger12}
Daniel Mellinger, Alex Kushleyev, and Vijay Kumar.
\newblock Mixed-integer quadratic program trajectory generation for
  heterogeneous quadrotor teams.
\newblock In {\em IEEE International Conference on Robotics and Automation},
  pages 477--483, 2012.

\bibitem{augugliaro12}
Federico Augugliaro, Angela~P. Schoellig, and Raffaello D'Andrea.
\newblock Generation of collision-free trajectories for a quadrocopter fleet: A
  sequential convex programming approach.
\newblock In {\em IEEE/RSJ International Conference on Intelligent Robots and
  Systems}, pages 1917--1922, 2012.

\bibitem{lavalle01}
Steven~M. LaValle and James~J. Kuffner.
\newblock Randomized kinodynamic planning.
\newblock {\em The International Journal of Robotics Research}, 20(5):378--400,
  2001.

\bibitem{khatib86}
Oussama Khatib.
\newblock Real-time obstacle avoidance for manipulators and mobile robots.
\newblock {\em The International Journal of Robotics Research}, 5(1):90--98,
  1986.

\bibitem{fiorini98}
Paolo Fiorini and Zvi Shiller.
\newblock Motion planning in dynamic environments using velocity obstacles.
\newblock {\em The International Journal of Robotics Research}, 17(7):760--772,
  1998.

\bibitem{alonsomora13icra}
Javier Alonso-Mora, Martin Rufli, Roland Siegwart, and Paul Beardsley.
\newblock Collision avoidance for multiple agents with joint utility
  maximization.
\newblock In {\em IEEE International Conference on Robotics and Automation},
  2013.

\bibitem{nate13}
Nate Derbinsky, Jos\'e Bento, Veit Elser, and Jonathan~S. Yedidia.
\newblock An improved three-weight message-passing algorithm.
\newblock {\em arXiv:1305.1961 [cs.AI]}, 2013.

\bibitem{Forney}
G.~David~Forney Jr.
\newblock Codes on graphs: Normal realizations.
\newblock {\em IEEE Transactions on Information Theory}, 47(2):520--548, 2001.

\bibitem{karaman2010incremental}
Sertac Karaman and Emilio Frazzoli.
\newblock Incremental sampling-based algorithms for optimal motion planning.
\newblock {\em arXiv preprint arXiv:1005.0416}, 2010.

\bibitem{glowinski75}
R.~Glowinski and A.~Marrocco.
\newblock Sur l'approximation, par \'el\'ements finis d'ordre un, et la
  r\'esolution, par p\'enalisization-dualit\'e, d'une class de probl\`ems de
  {D}irichlet non lin\'eare.
\newblock {\em Revue Fran\c{c}aise d'Automatique, Informatique, et Recherche
  Op\'erationelle}, 9(2):41--76, 1975.

\bibitem{gabay76}
Daniel Gabay and Bertrand Mercier.
\newblock A dual algorithm for the solution of nonlinear variational problems
  via finite element approximation.
\newblock {\em Computers \& Mathematics with Applications}, 2(1):17--40, 1976.

\bibitem{everett63}
Hugh~Everett III.
\newblock Generalized lagrange multiplier method for solving problems of
  optimum allocation of resources.
\newblock {\em Operations Research}, 11(3):399--417, 1963.

\bibitem{hestenes69a}
Magnus~R. Hestenes.
\newblock Multiplier and gradient methods.
\newblock {\em Journal of Optimization Theory and Applications}, 4(5):303--320,
  1969.

\bibitem{hestenes69b}
Magnus~R. Hestenes.
\newblock Multiplier and gradient methods.
\newblock In L.A.~Zadeh et~al., editor, {\em Computing Methods in Optimization
  Problems 2}. Academic Press, New York, 1969.

\bibitem{powell69}
M.J.D. Powell.
\newblock A method for nonlinear constraints in minimization problems.
\newblock In R.~Fletcher, editor, {\em Optimization}. Academic Press, London,
  1969.

\bibitem{boyd10}
Stephen Boyd, Neal Parikh, Eric Chu, Borja Peleato, and Jonathan Eckstein.
\newblock Distributed optimization and statistical learning via the alternating
  direction method of multipliers.
\newblock {\em Foundations and Trends in Machine Learning}, 3(1):1--122, 2011.

\end{thebibliography}

\newpage
\section*{ \Large Supplementary material for ``A message-passing algorithm for multi-agent trajectory planning''}

This document gives details
on the message-passing algorithm we use
and how to implement all minimizers described in
the paper. It also includes some more comments on
our numerical results and related literature.
A short movie clip showing the behaviour of our algorithm can be found at {\color{blue} \emph{http://youtu.be/yuGCkVT8Bew}}

\appendix

%
%

\section{Comment on related literature} 
\label{app:more_on_literature}

A$^*$-search based methods and sampling-based methods require exploring a continuous domain using discrete graph structures. For problems with many degrees of freedom or complex kinematic and dynamic constraints, as when dealing with multiple agents or manipulators, fixed-grid search methods are impractical. Alternatively, exploration can be done using sampling algorithms with proved asymptotic convergence to the optimal solution \cite{karaman2010incremental}. However, as the dimensionality of the configuration space increases, the convergence rate degrades and the local planners required by the exploration loop become harder to implement. In addition, as pointed out in \cite{lavalle01}, sampling algorithms cannot easily produce solutions where multiple agents move in tight spaces, like in CONF1 with obstacles. Some of the disadvantages of using discrete random search structures are even visible in extremely simple scenarios. For example, for�\, a single holonomic agent that needs to move as quickly as possible between two points in free-space, \cite{karaman2010incremental} require around 10000 samples on their RRT* method to find something close to the shortest-path solution. For our algorithm this is a trivial scenario: it outputs the optimal straight-line solution in 200 iterations and 37 msecs. in our Java implementation.

%
%

\section{Full description of the improved three-weight  message-passing algorithm of \cite{nate13}} 
\label{app:mod_mess_pass_admm}

First we give a self-contained
(complete) description of the three-weight algorithm TWA from \cite{nate13}. Their method is an improvement of the alternating direction method of multipliers (ADMM) \footnote{ADMM is a decomposition procedure for solving optimization problems. It coordinates the solutions to small local sub-problems to solve a large global problem. Hence, it is useful to derive parallel algorithms. It was introduced in \cite{glowinski75} and \cite{gabay76} but is closely related to previous work as the dual decomposition method introduced by \cite{everett63} and the method of multipliers introduced by \cite{hestenes69a,hestenes69b} and \cite{powell69}.
For a good review on ADMM see \cite{boyd10}, where you can also find a self-contained proof of its convergence for convex problems.}.
Assume we want to solve
\begin{equation} \label{eq:main_opt_problem}
\min_{x \in \reals^d} \sum^l_{b = 1} f_b(x_{\partial b}),
\end{equation}
where the set $x_{\partial b} = \{x_j: j \in  \partial b\}$ is a vector obtained by considering the subset of entries of $x$ with index in $\partial b \subseteq [d]$. The functions $f_b$ do not need to be convex or smooth for the algorithm to be well-defined. However, the algorithm is only guaranteed to find the global minimum under convexity \cite{boyd10}.

\begin{figure}[h!]
\begin{center}
\includegraphics[width=0.3\textwidth]{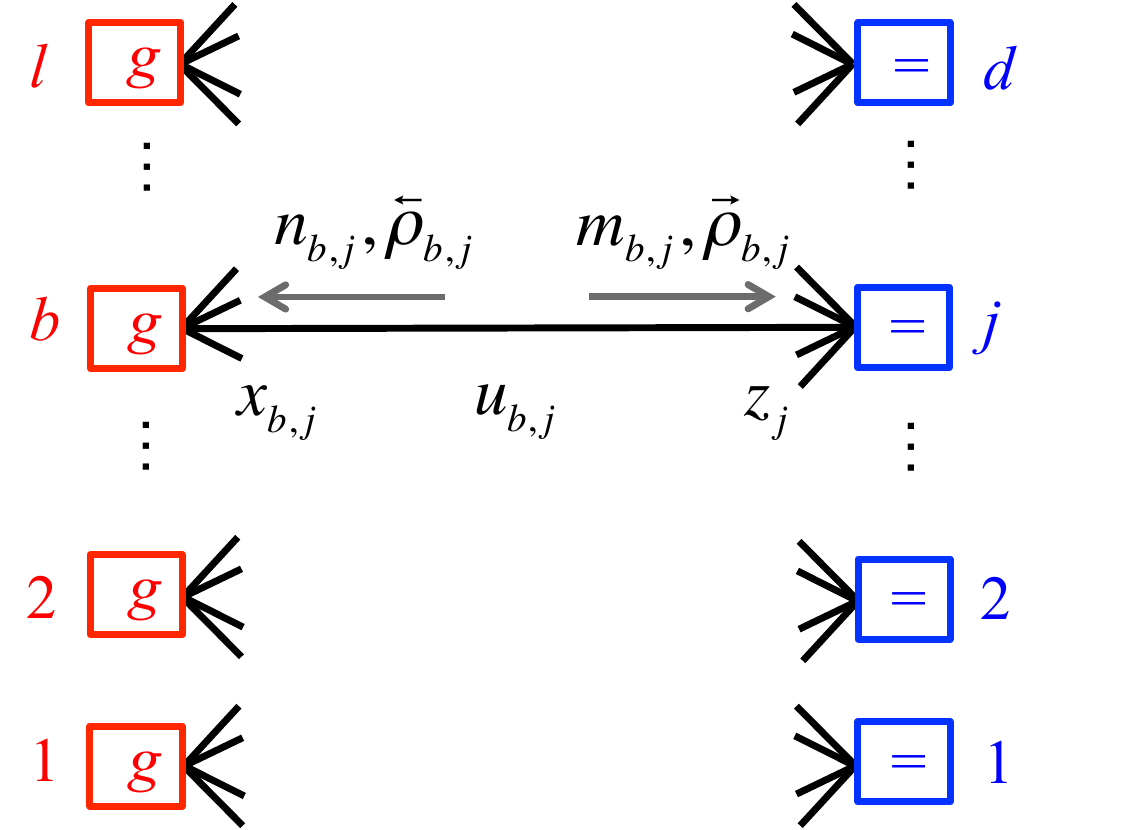}
\caption{Bipartite graph used by the message-passing algorithm to solve \eqref{eq:main_opt_problem}. The algorithm works by updating seven kind of variables, also shown in picture.}
\end{center}
\label{fig:bipartite_general_case}
\end{figure}

Start by forming the following bipartite graph consisting of minimizer-nodes and equality-nodes.
Create one minimizer-node, labeled ``$g$'', per function $f_b$ and one equality-node, labeled ``='', per variable $x_j$. There are $l$ minimizer-nodes and $d$ equality-nodes in total.
If function $f_b$ depends on variable $x_j$,  create an edge $(b,j)$ connecting $b$ and $j$ (see Figure \ref{fig:bipartite_general_case} for a general representation).

The algorithm in \cite{nate13} works by repetitively updating seven kind of variables. These can be listed as follows.
Every equality-node $j$ has a corresponding variable $z_j$. Every edge $(b,j)$ from minimizer-node $b$ to equality-node $j$ has a corresponding variable $x_{b,j}$ , variable $u_{b,j}$, message $n_{b,j}$, message $m_{b,j}$, weight $\overrightarrow{\rho}_{b,j}$ and  weight $\overleftarrow{\rho}_{b,j}$.

To start the method, one specifies the initial values $\{z^0_j\}$, 
$\{u^0_{b,j}\}$ and $\{\overleftarrow{\rho}^0_{b,j}\}$. Then, at every iteration $k$, repeat the following.
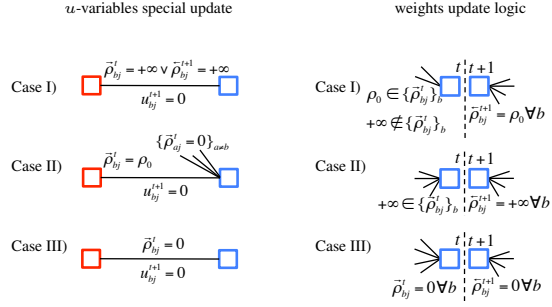
\begin{figure}[t]
\begin{center}
\includegraphics[width=0.45\textwidth]{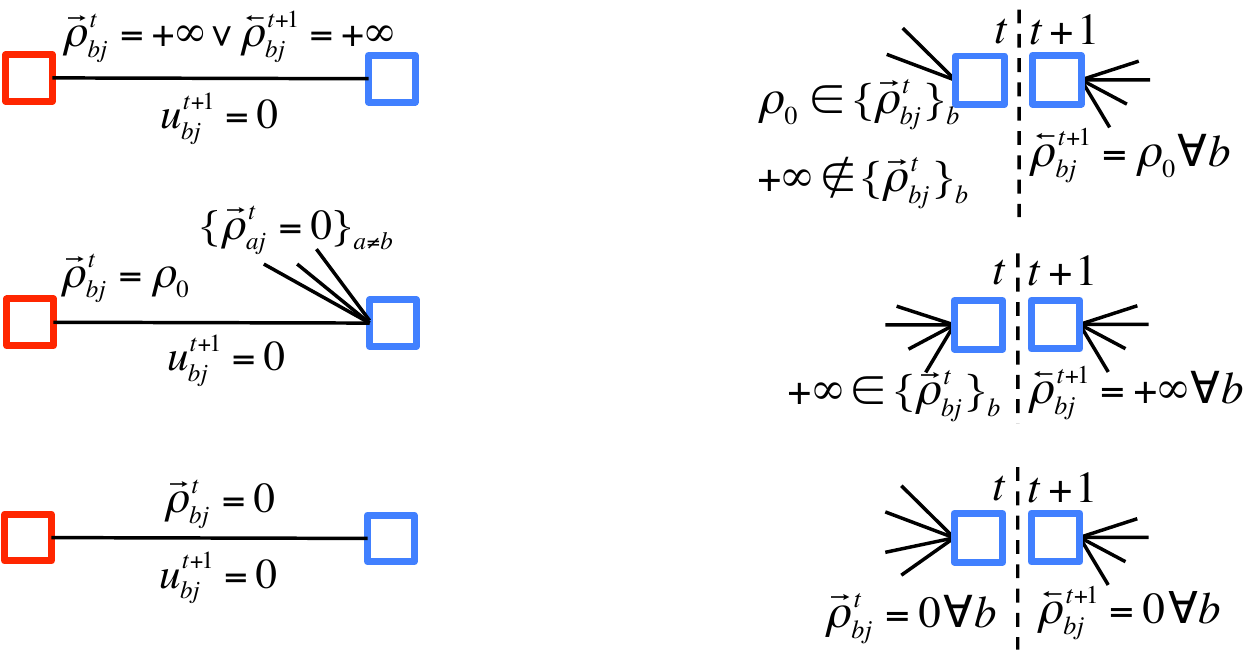}
\put(-60,110){\tiny weights update logic}
\put(-185,110){\tiny $u$-variables special update}
\put(-205,80){\tiny Case I)}
\put(-205,50){\tiny Case II)}
\put(-205,20){\tiny Case III)}
\put(-90,80){\tiny Case I)}
\put(-90,50){\tiny Case II)}
\put(-90,20){\tiny Case III)}
\caption{Left: Special update-rules for variables $u$. Right: Update rule for variables $\protect\overleftarrow{\rho}$.}
\label{fig:special_update_rules}
\end{center}
\end{figure}
\begin{enumerate}
\item Construct the $n$-message for every edge $(b,j)$ as $n^k_{b,j} = z^k_{j} - u^k_{b,j}$.\\
\item Update the $x$-variables for every edge $(b,j)$. All $x$-variables associated to edges incident on the same minimizer-node $b$, i.e. $\{x^k_{b,j}\}_{j \in \partial b}$, are updated simultaneously by the minimizer $g_b$ associated to the function $f_b$,
\begin{align*}
\{x^{k+1}_{b,j}\}_j &= g_b \left(\{n^{k}_{b,j}\}_j,\{\overleftarrow{\rho}^{k}_{b,j}\}_j\right)\nonumber\\
&\equiv \arg \min_{ \{\tilde{x}^k_{b,j}\}_j  } f_b( \{\tilde{x}^k_{b,j}\}_j   ) +\frac{1}{2} \sum_j \overleftarrow{\rho}^k_{b,j} (\tilde{x}^k_{b,j} - n^k_{b,j})^2.
\end{align*}
In the expression above we write $\{ \}_j$ and $\sum_j$ for $\{ \}_{j \in \partial b}$ and $\sum_{j \in \partial b}$ respectively. If the $\arg \min$ returns a set with more than one element, choose a value uniformly at random from this set.\\
\item Compute the outgoing weights for all edges
$(b,j)$. All weights associated to edges leaving the same minimizer-node $b$,
i.e. $\{\overrightarrow{\rho}^k_{b,j}\}_{j \in \partial b}$, are updated simultaneously
according to a user-defined logic that should depend on the problem defined by equation \eqref{eq:main_opt_problem}. This logic can assign three possible values for the weights: $\overrightarrow{\rho}^k_{b,j} = 0$,
$\overrightarrow{\rho}^k_{b,j} = \rho_0$ or $\overrightarrow{\rho}^k_{b,j} = \infty$, where $\rho_0 > 0$ is some pre-specified constant. The idea is to use these three values to inform the equality-nodes of how certain the minimizer-node $b$ is that the current variable $x^k_{b,j}$ is the optimal value for variable $x_j$ in equation \eqref{eq:main_opt_problem}. A weight of $\infty$ is used to signal total certainty, $0$ for no certainty at all and $\rho_0$ for all the other scenarios. Later, equality-nodes average the information coming from the minimizer-nodes by these weights to update the consensus variables $z$.\\
\item Construct the $m$-messages for every edge $(b,j)$ as 
$m^k_{b,j} = x^{k+1}_{b,j} + u^k_{b,j}$.\\
\item  Update the $z$-variable for
every equality-node $j$ as $$z^{k+1}_j =
\frac{\sum_{b \in \partial j}  \overrightarrow{\rho}^k_{b,j} m^k_{b,j}}{ \sum_{b \in \partial j} \overrightarrow{\rho}^k_{b,j}}.$$
The set $\partial j$ contains all the minimizer-nodes that connect to equality-node $j$. If all weights $\{\overrightarrow{\rho}^k_{b,j}\}_{b \in \partial j}$ are zero in the previous expression, treat them as $1$.\\
\item Compute the updated weights $\overleftarrow{\rho}^{k+1}$. Weights leaving the same equality-node $j$, i.e.$\{\overleftarrow{\rho}^{k+1}_{b,j}\}_{b \in \partial j}$, are computed simultaneously. The update logic is described in Figure \ref{fig:special_update_rules}-right. According to which of the three distinct scenarios $\overrightarrow{\rho}^{k}$ falls, $\overleftarrow{\rho}^{k+1}$ is uniquely determined. This logic again assigns three possible values for the weights $\{0, \rho_0, \infty \}$.\\
\item Update the $u$-variables for all edges. If edge $(b,j)$ has $\overrightarrow{\rho}^k_{b,j} = \rho_0$ and $\overleftarrow{\rho}^{k+1}_{b,j} = \rho_0$ then $u^{k+1}_{b,j} = u^{k}_{b,j} + (\alpha/\rho_0) (x^{k+1}_{b,j} - z^{k+1}_j)$, where $\alpha$ is a pre-specified constant. Otherwise, choose $u^{k+1}_{b,j}$ according to the three scenarios described in Figure \ref{fig:special_update_rules}-left, depending on the weights $\overrightarrow{\rho}$ and $\overleftarrow{\rho}$.
\end{enumerate}

In short, given $\rho_0$, $\alpha$, the initial values $\{z^0_j\}$, 
$\{u^0_{b,j}\}$ and $\{\overleftarrow{\rho}^0_{b,j}\}$, all the minimizers $\{g_b\}$, with corresponding update logic for $\overrightarrow{\rho}$, and the bipartite graph, the method is completely specified. If all weights, $\overrightarrow{\rho}$ and $\overleftarrow{\rho}$, are set to $\rho_0$ across all iterations, the described method reduces to classical ADMM , interpreted as a message-passing algorithm. Finally, notice that at each time step $k$, all the variables associated with each edge can be updated in parallel. In particular, the update of the $x$-variables, usually the most expensive operation, can be parallelized.

%
%

\section{Agent-agent collision minimizer} \label{app:agent_agent_minimizer}

Here we give the details of how to write the agent-agent
collision minimizer, $g^{\text{coll}}$ using the
agent-obstacle minimizer $g^{\text{wall}}$ of equation \eqref{eq:wall_minimizer_problem}.

First recall that $f^{\text{coll}}_{r,r'}(\ux,\ox,\ux',\ox') = f^{\text{wall}}_{0,0,r+r'}(\ux'-\ux,\ox'-\ox)$. Then, rewrite \eqref{eq:coll_minimizer_problem} as,
\begin{align} \label{eq:coll_to_wall_step1}
&g^{\text{coll}}(\un,\on,\un',\on',\urho,\orho,\urho',\orho',r,r') =  \arg \min_{\{\ux,\ox,\ux',\ox'\}} 
\bigg[ f^{\text{wall}}_{0,0,r+r'}(\ux'-\ux,\ox'-\ox)\nonumber\\
& + \frac{\urho}{2} \|\ux - \un \|^2 + \frac{\orho}{2} \|\ox - \on \|^2 + \frac{\urho'}{2} \|\ux' - \un' \|^2 + \frac{\orho'}{2} \|\ox' - \on' \|^2
\bigg] .
\end{align}
Now introduce the following variables $\uv= \ux' - \ux$, $\uu = \ux'+\ux$, $\ov = \ox' - \ox$ and $\ou = \ox' + \ox$. The function being minimized in \eqref{eq:coll_to_wall_step1} can be written as
\begin{align} \label{eq:eq:coll_to_wall_step2}
&f^{\text{wall}}_{0,0,r+r'}(\uv,\ov)
+ \frac{\urho}{2} \Big\|\frac{\uu-\uv}{2} - \un \Big\|^2 + \frac{\orho}{2} \Big\|\frac{\ou-\ov}{2} - \on \Big\|^2 \nonumber\\
& + \frac{\urho'}{2} \Big\|\frac{\uu+\uv}{2} - \un' \Big\|^2 + \frac{\orho'}{2} \Big\|\frac{\ou+\ov}{2} - \on' \Big\|^2.
\end{align}
Now notice that we can write,
\begin{align*}
& \frac{\urho}{2} \Big\|\frac{\uu-\uv}{2} - \un \Big\|^2  + \frac{\urho'}{2} \Big\|\frac{\uu+\uv}{2} - \un' \Big\|^2 = \frac{\urho}{8} \|\uu-\uv - 2\un \|^2  + \frac{\urho'}{8} \|\uu+\uv - 2 \un' \|^2\\
& = \frac{\urho}{8} (\|\uu\|^2 + \| \uv + 2\un \|^2  - 2 \langle \uu, \uv+2\un\rangle)+ \frac{\urho'}{8} (\|\uu\|^2 + \|\uv - 2 \un' \|^2 + 2 \langle \uu,\uv-2\un'\rangle)\\
& = \frac{\urho+\urho'}{8} \|\uu\|^2 +  2 \Big \langle \uu, \frac{\urho'-\urho}{8} \uv-\frac{\urho \un +\urho' \un'}{4} \Big\rangle  + \frac{\urho}{8}\| \uv + 2\un \|^2 + \frac{\urho'}{8}\|\uv - 2 \un' \|^2\\
& = \frac{\urho + \urho'}{8}\Big\|\uu - \left(\frac{\urho-\urho'}{\urho+\urho'} \uv+\frac{2(\urho \un +\urho' \un')}{\urho+\urho'} \right) \Big \|^2 + \frac{\urho + \urho'}{8}\Big\|\uv - \frac{2(\urho' \un' -\urho \un)}{\urho+\urho'}  \Big \|^2\\
&+C(\un,\urho,\un',\urho'),
\end{align*}
where $C(\un,\urho,\un',\urho')$ is a constant that depends on the variables $\{\un,\urho,\un',\urho'\}$.
A similar manipulation can be done to $ \frac{\orho}{2} \Big\|\frac{\ou-\ov}{2} - \on \Big\|^2 + \frac{\orho'}{2} \Big\|\frac{\ou+\ov}{2} - \on' \Big\|^2$. 
Therefore, the expression \eqref{eq:eq:coll_to_wall_step2}
can be rewritten as
\begin{align} \label{eq:eq:coll_to_wall_step3}
&f^{\text{wall}}_{0,0,r+r'}(\uv,\ov)
+ \frac{\urho+\urho'}{8} \Big\| \uv - \frac{2(\urho' \un' - \urho \un)}{\urho+\urho'}   \Big\|^2 +
\frac{\orho+\orho'}{8} \Big\| \ov - \frac{2(\orho' \on' - \orho \on)}{\orho+\orho'}   \Big\|^2 \nonumber\\
&+ \frac{\urho+\urho'}{8} \Big\| \uu - \left(  \frac{\urho - \urho'}{\urho  + \urho'} \uv + \frac{2(\urho \un + \urho' \un')}{\urho + \urho'}  \right)  \Big\|^2 \nonumber\\
& +  \frac{\orho+\orho'}{8} \Big\| \ou - \left(  \frac{\orho - \orho'}{\orho  + \orho'} \ov + \frac{2(\orho \on + \orho' \on')}{\orho + \orho'}  \right)  \Big\|^2+ C(\un,\un',\on,\on',\urho,\urho',\orho,\orho'),
\end{align}
where $C(\un,\un',\on,\on',\urho,\urho',\orho,\orho')$ is a constant that depends on the variables
$\un$, $\un'$, $\on$, $\on'$, $\urho$, $\urho'$, $\orho$ and
$\orho'$.

Let $\{\uv^*,\ov^*,\uu^*,\ou^*\}$ be a set of values
that minimizes equation \eqref{eq:eq:coll_to_wall_step3}.
We have,
\begin{align}
\{\uv^*,\ov^*\} &\in g^{\text{wall}}\left(\frac{2(\urho' \un'-\urho \un)}{\urho+\urho'},\frac{2(\orho' \on'- \orho \on)}{\orho+\orho'},r+r',0,0,\frac{\urho+\urho'}{4},\frac{\orho+\orho'}{4}\right), \label{eq:linear_tranformation_wall_coll_1}\\
\{\uu^*,\ou^* \} &= \Big \{\frac{\urho - \urho'}{\urho  + \urho'} \uv^* + \frac{2(\urho \un + \urho' \un')}{\urho + \urho'} ,
\frac{\orho - \orho'}{\orho  + \orho'} \ov^* + \frac{2(\orho \on + \orho' \on')}{\orho + \orho'} \Big\} \label{eq:linear_tranformation_wall_coll_2}.
\end{align}
We can now produce a set of values that satisfy
$$\{\ux^*,\ox^*,\ux'^*,\ox'^*\} \in g^{\text{coll}}(\un,\on,\un',\on',\urho,\orho,\urho',\orho',r,r')$$
using the following relation,
\begin{align*}
\{\ux^*,\ox^*,\ux'^*,\ox'^*\} &= \Big \{\frac{\uu^* - \uv^*}{2},\frac{\ou^* - \ov^*}{2}, \frac{\uv^*+\uu^*}{2},\frac{\ou^* + \ov^*}{2} \Big \}.
\end{align*}
In fact, all values $\{\ux^*,\ox^*,\ux'^*,\ox'^*\} \in g^{\text{coll}}(\un,\on,\un',\on',\urho,\orho,\urho',\orho',r,r')$ can be obtained from some
$\{\uv^*,\ov^*,\uu^*,\ou^*\}$ that minimizes equation
\eqref{eq:eq:coll_to_wall_step3}.
In other words, the minimizer $g^{\text{coll}}$ can be
expressed in terms of the minimizer $g^{\text{wall}}$
by means of a linear transformation.

Minimizers can receive zero-weight messages $\overleftarrow{\rho}$ from
their neighboring equality nodes.
In \eqref{eq:linear_tranformation_wall_coll_1} and \eqref{eq:linear_tranformation_wall_coll_2},
this can lead to indeterminacies. We address this as follows.
If $\urho$ and $\urho'$ are simultaneously zero then we compute
\eqref{eq:linear_tranformation_wall_coll_1} and \eqref{eq:linear_tranformation_wall_coll_2}
in the limit when $\urho = \urho' \rightarrow 0^+$. When implementing this on
software, we simply replace them by small equal values. The fact that $\urho = \urho'$
resolves the indeterminacies in the fractions
and taking the limit to zero from above guarantees that
the wall minimizer $g^{\text{wall}}$, that is solved using a mechanical analogy involving springs,
is well behaved (See Section \ref{app:mechanical_analog}).
If $\orho$ and $\orho'$ are simultaneously zero, we perform a similar operation.

%
%

\section{Agent-obstacle collision minimizer} \label{app:agent_obstacle_minimizer}

In Section \ref{app:agent_agent_minimizer}
we expressed the agent-agent collision
minimizer by applying a linear
transformation to the agent-obstacle collision minimizer.
Now we show how the agent-obstacle minimizer can
be posed as a classical mechanical problem involving a system
of springs. Although the relationship in Section \ref{app:agent_agent_minimizer} holds in general, the transformation presented in this section holds only when the agents move in the plane, i.e. $x_i(s) \in \reals^2 \;\;\forall s,i$. Similar transformations should hold in higher dimensions.

When the obstacle is a line-segment $[x_L, x_R]$, the agent-obstacle minimizer \eqref{eq:wall_minimizer_problem} solves the following non-convex optimization problem,
\begin{align}
&\underset{\{\ux,\ox\}}{\text{minimize }} \;\; \bigg[ \frac{\urho}{2} \|\ux -\un \|^2 + \frac{\orho}{2} \|\ox -\on \|^2
\bigg]
\label{eq:wall_min_explicit_opt_problem}\\
&\text{s.t. } \|(\alpha \ox + (1 - \alpha) \ux) - (\beta x_R + (1 - \beta) x_L)\| \geq r \text{ for all } \alpha, \beta \in [0,1] \label{eq:wall_minimizer_line_constraint}.
\end{align}

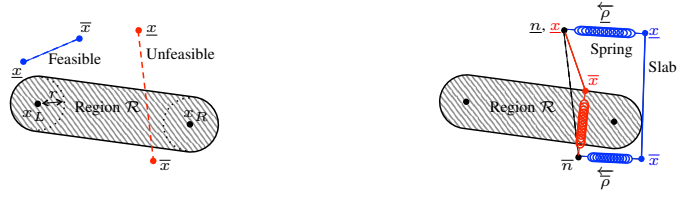
\begin{figure}[h!]
\begin{center}
\includegraphics[width=0.2\textwidth]{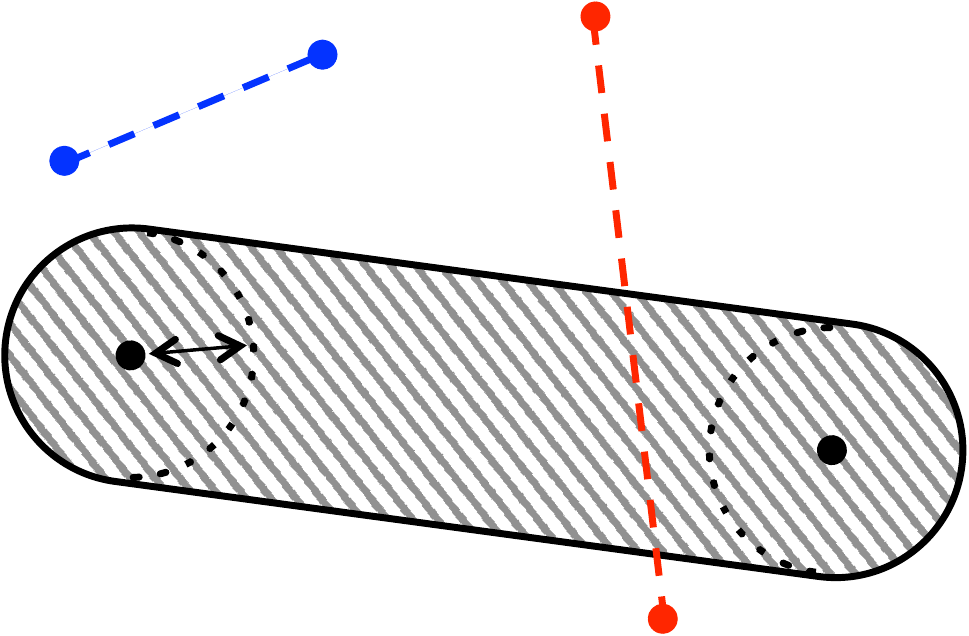}
\;\;\;\;\;\;\;\;\;\;\;\;\;\;\;\;\;\;\;\;\;\;\;\;\;\;\;\;
\includegraphics[width=0.2\textwidth]{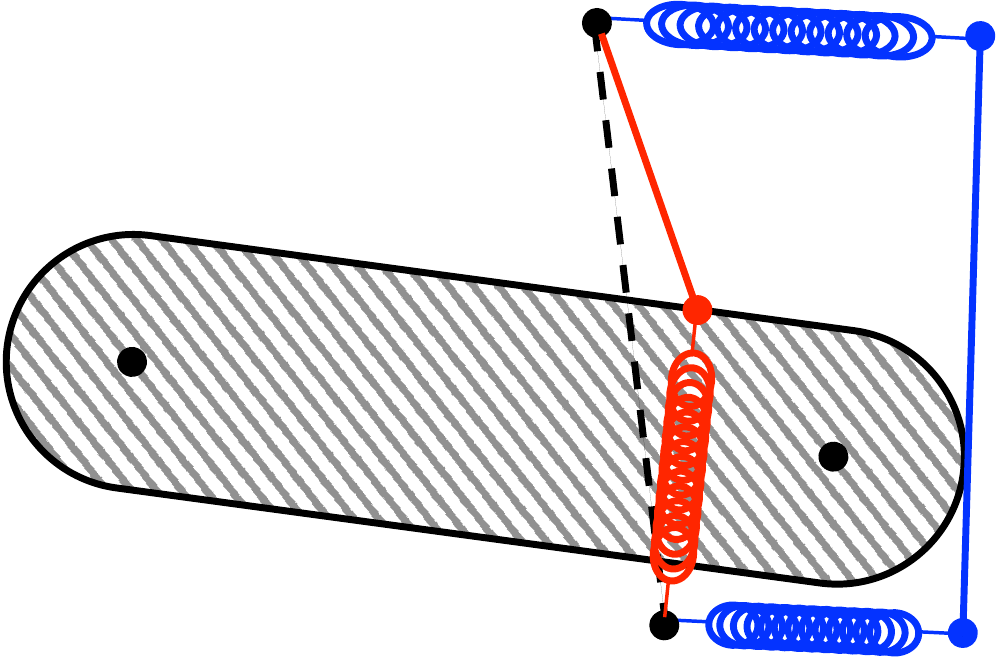}
\put(-0,35){\tiny Slab}
\put(-22,42){\tiny Spring}
\put(-60,20){\tiny Region $\mathcal{R}$}
\put(-0,49){\tiny \color{blue} \ux}
\put(-23,30){\tiny \color{red} \ox} %
\put(-0,00){\tiny \color{blue} \ox}
\put(-20,56){\tiny \urho}
\put(-20,-8){\tiny \orho}
\put(-45,51){\tiny \un, \color{red} \ux }
\put(-33,-2){\tiny \on}
\put(-217,20){\tiny Region $\mathcal{R}$}
\put(-227,38){\tiny Feasible}
\put(-190,40){\tiny Unfeasible}
\put(-237,18){\tiny $x_L$}
\put(-227,25){\tiny $r$}
\put(-176,18){\tiny $x_R$}
\put(-241,34){\tiny $\ux$}
\put(-215,50){\tiny $\ox$}
\put(-190,50){\tiny $\ux$}
\put(-185,-2){\tiny $\ox$}
\caption{Left: Feasible solution (blue) and unfeasible solution (red). Right:  Two different feasible configurations of the springs-slab system. Each represented in different color.}
\label{fig:mechanical_region}
\end{center}
\end{figure}

Observe that the term $\frac{\urho}{2} \|\ux -\un \|^2$ equals the energy of a spring with zero rest-length and elastic coefficient $\urho$ whose end points are at positions $\ux$ and $\un$. The same interpretation applies for the second term in \eqref{eq:wall_min_explicit_opt_problem}. With this interpretation in mind, the non-convex constraint \eqref{eq:wall_minimizer_line_constraint} means that
the line from $\ux$ to $\ox$ cannot cross the region
swept out by a circle of radius $r$ that moves from $x_L$ to $x_R$. We call this region $\mathcal{R}$. Figure \ref{fig:mechanical_region}-left shows a
feasible solution and an unfeasible solution under this interpretation.
When the line from $\un$ to $\on$ does not
cross $\mathcal{R}$, the solution of \eqref{eq:wall_min_explicit_opt_problem}-\eqref{eq:wall_minimizer_line_constraint} is $\ux = \un$
and $\ox = \un$. In general however, $\ux$ and $\ox$ adopt
the minimum energy configuration of a system with two zero rest-length springs, with end points $(\un,\ux)$ and $(\on,\ox)$ and elastic coefficients $\urho$ and $\orho$, and with a hard extensible slab, connecting $\ux$ to $\ox$, that
cannot go over region $\mathcal{R}$. The slab can
be extended without spending any energy.
Figure \ref{fig:mechanical_region}-right shows two
feasible configurations of the system of springs and slab
when $\ux=\un$ and $\ox = \on$ cannot be a feasible solution.

It is possible that this minimizer receives two zero-weight messages from its neighboring
equality nodes, i.e., $\urho = \orho = 0$. This would correspond to not having
any spring connecting point $\un$ to $\ux$ and $\on$ to $\ox$. The mechanic
system would then be indeterminate. When this is the case, we solve the mechanic system
in the limit when $\urho = \orho \rightarrow 0^+$. In terms of software implementation, this
is achieved by replacing $\urho$ and $\orho$ by small equal values.

In Section \ref{app:mechanical_analog}
we explain how to compute the
minimum energy configuration of this system
quickly. In other words, we show that the minimizer
$g^{\text{wall}}$ can be implemented efficiently.

%
%

\section{Energy minimizer} \label{app:energy_minimizer}

The energy minimizer solves the quadratic
optimization problem 
$$\min_{\{\ux, \ox\}} \bigg[ C \| \ux - \ox\|^2 + (\urho/2) \|\ux - \un \|^2 + (\orho/2) \|\ox - \on \|^2\bigg].$$ 
From the first order optimality conditions we get
$2C (\ux - \ox) + \urho (\ux - \un) = 0$ and
$2C (\ox - \ux) + \orho (\ox - \on) = 0$.
Solving for $\ux$ and $\ox$ we obtain,
\begin{equation} \label{eq:energy_minimizer_output}
\ux = \frac{\urho \orho \un + 2C(\urho \un + \orho \on)}{2C (\urho + \orho) + \urho \orho},\;\;\;
\ox = \frac{\urho \orho \on + 2C(\urho \un + \orho \on)}{2C (\urho + \orho) + \urho \orho}.
\end{equation}

If the energy minimizer receives $\urho = \orho = 0$, we resolve the indeterminacy in computing
\eqref{eq:energy_minimizer_output} by letting $\urho = \orho \rightarrow 0^+$.

%
%

\section{Maximum velocity minimizer} \label{app:max_vel_minimizer}

This minimizer solves the convex problem $\text{minimize}_{\{\ux,\ox\}} \bigg[ (\urho/2) \|\ux - \un \|^2 + (\orho/2) \|\ox - \on \|^2 \bigg]$ subject to $\| \ux - \ox\| \leq C$. If $\|\un - \on \| \leq C$ then $\ux = \un$ and $\ox = \on$. Otherwise, the constraint is active, and, using the KKT conditions, we have $\urho (\ux - \un) = -\lambda (\ux - \ox)$ and $\orho (\ox - \on) = -\lambda (\ox - \ux)$ where
$\lambda \neq 0$ is such that $\|\ux - \ox \| = C$. Solving for $\ux$ and $\ox$ we get,
\begin{align}\label{eq:sol_paramterized_max_vel}
\ux = \frac{\urho (\orho + \lambda) \un + \lambda \orho \on}{\urho \orho + \lambda (\orho  + \urho)}, \;\;\;
\ox = \frac{\orho (\urho + \lambda) \on + \lambda \urho \un}{\urho \orho + \lambda (\orho  + \urho)}.
\end{align}
To find the solution we just need to determine $\lambda$.
Computing the difference between the above expressions we get,
\begin{equation} \label{eq:sol_paramterized_max_vel_diff}
\ux - \ox = \frac{\un - \on}{1 + (\frac{1}{\urho} + \frac{1}{\orho})\lambda}.
\end{equation}
Taking the norm of the right hand side and setting it equal to $C$ we get
\begin{equation} \label{eq:lambda_val_max_vel}
\lambda = \pm \frac{ (\|\un - \on \|/C) - 1}{\urho^{-1} + \orho^{-1}}.
\end{equation}

Now examine equation \eqref{eq:sol_paramterized_max_vel_diff}. Starting from an $\un - \on$ such
that $ \|\un - \on \| > C$, the fastest way to get to $\ux - \ox$ with $\|\ux - \ox\| = C$ is to increase
$\lambda > 0$. Hence, in \eqref{eq:lambda_val_max_vel}, we should choose the positive solution, i.e. 
\begin{equation} \label{eq:unique_lambda_val_max_vel}
\lambda = \frac{ (\|\un - \on \|/C) - 1}{\urho^{-1} + \orho^{-1}}.
\end{equation}

If the maximum velocity minimizer receives $\urho = \orho = 0$, we resolve any indeterminacy
by letting $\urho = \orho \rightarrow 0^+$. In software, this is achieved by setting $\urho$ equal to $\orho$
equal to some small value.
%
%

\section{Minimum velocity minimizer} \label{app:min_vel_minimizer}

This minimizer can be computed in a very similar
way to the maximum velocity minimizer.
If $\|\un - \on \| \geq C$, then $\ux = \un$ and $\ox = \on$.
Otherwise, from the KKT conditions, we again obtain equation
\eqref{eq:sol_paramterized_max_vel}. The difference
$\ux - \ox$ is again the expression \eqref{eq:sol_paramterized_max_vel_diff}. Now, however,
starting from $\un - \on$ such that $\|\un - \on \|>C$, the fastest way to get 
to $\ux - \ox$ with $\|\ux - \ox\| = C$, is to decrease
$\lambda < 0$. Hence, in \eqref{eq:lambda_val_max_vel}, we should choose the negative solution, i.e. \eqref{eq:unique_lambda_val_max_vel} holds again. If the minimum velocity minimizer receives $\urho = \orho = 0$, we resolve any indeterminacy
by letting $\urho = \orho \rightarrow 0^+$. In software, this is achieved by setting $\urho$ equal to $\orho$
equal to some small value.

%
%

\section{Velocity obstacle minimizers} \label{app:VO_minimizer}

In this section we explain how to write the minimizers associated to each of the terms in
equation \eqref{eq:unconst_opt_RVO} using the minimizers $g^{\text{coll}}$, $g^{\text{wall}}$ and $g^{\text{cost}}$
for global planning.

First however, we briefly describe the bipartite graph that
connects all these minimizers together. The bipartite graph for this problem has a $g^{\text{VO coll}}$ minimizer-node
connecting every pair of equality-nodes. There is one equality-node per variable in $\{x_i\}$. Recall that each of these variables describes the position of an agent at the end of a planning epoch. Each equality-node is also connected to a separate $g^{\text{VO cost}}$ minimizer-node. Finally, for obstacles in 2D, every equality node is also connected to several $g^{\text{VO wall}}$ minimizer-nodes, one per obstacle.

We start by describing the minimizer associated to the terms $\{f^{\text{coll}}_{r_i,r_j} (x_i(0),x_i,x_j(0),x_j) \}$ in equation \eqref{eq:unconst_opt_RVO}. This is given by
\begin{equation*}
g^{\text{VO coll}}(\un,\un',\urho,\urho',x^0,x'^0,r,r') = 
\arg \min_{\{\ux,\ux'\}} \bigg[ f^{\text{coll}}_{r,r'} (x^0,\ux,x'^0,\ux')
+ \frac{\urho}{2}  \|\ux - \un \|^2
+ \frac{\urho'}{2}  \|\ux' - \un' \|^2 \bigg].
\end{equation*}
The messages $\un$ and $\un'$, and corresponding certainty weights $\urho$ and $\urho'$, come from
the equality-nodes associated to the end position
of two agents of radius $r$ and $r'$ that, during one time epoch, move from their initial positions $x^0$ and $x'^0$ to $\ux$ and $\ux'$ without colliding.

The outgoing weights $\overrightarrow{\underline{\rho}}$ and 
$\overrightarrow{\underline{\rho}'}$ associated to the variables
$\ux$ and $\ux'$ are determined in the following way. If an agent of radius $r$ moving from $x^0$ to $\un$ does not collide with an agent of radius $r'$ moving from $x'^0$ to
$\un'$, the minimizer will not propose a new trajectory for them, i.e., the minimizer will return $\ux = \un$ and $\ux' = \un'$. Hence, in this case, we set all outgoing weights equal to $0$, signaling to neighboring equality-nodes that the minimizer wants to have no say when try to reach consensus. Otherwise, we set all outgoing weights equal to $\rho_0$.

For this minimizer, by direct substitution one sees that,
\begin{equation} \label{eq:minimizer_rvo_coll}
g^{\text{VO coll}}(\un,\un',\urho,\urho',x^0,x'^0,r,r') = g^{\text{coll}}(x^0,\un,x'^0,\un',+\infty,\urho,+\infty,\urho',r,r').
\end{equation}
Above we are using a notation where, given a function $f$, $f(+\infty) \equiv \lim_{x \rightarrow +\infty} f(x)$.
In software, this is implemented by replacing $+\infty$ by a very large value.

In a very similar way, the minimizer associated to the terms $\{f^{\text{wall}}_{{x_R}_k,{x_L}_k,r_i} (x_i(0),x_i)\}$
can be written using the agent-obstacle minimizer for the global planning problem. Concretely,
\begin{equation} \label{eq:minimizer_rvo_wall}
g^{\text{VO wall}}(\un,x_L,x_R,r,\urho) = g^{\text{wall}}(x^0,\un,r,x_L,x_R,+\infty,\urho).
\end{equation}
For this minimizer, the rule to set the outgoing weights is the following. If an agent of radius $r$ can move from $x^0$ to $\un$ without colliding with the line segment $[x_L \, x_R]$ then set all outgoing weights to $0$. Otherwise set them to $\rho_0$.

Finally, we turn to the the minimizer associated to the terms $\{f^{\text{cost}}_{C'_i} (x_i,x^{\text{ref}}_i )\}$. This minimizer 
receives as input a message $\un$, with corresponding certainty weight $\urho$, from the equality-minimizer associated to the position of an agent at the end of a time epoch and outputs a local estimate, $\ux$, of its position at the end of the epoch. It also receives as parameter a reference position $x^\text{ref}$ and a cost $c$ of $\ux$ deviating from it. To be concrete, its output is chosen uniformly at random from following $\arg \min$ set,
\begin{align} 
g^{\text{VO cost}}(\un,\urho,x^\text{ref},c) &= \arg \min_{\ux}
\bigg[ f^{\text{cost}}_{c} (\ux,x^{\text{ref}} ) + \frac{\urho}{2} \|\ux - \un \|^2 \bigg]\\
& = \arg \min_{\ux} \;\;
\bigg[ c \| \ux - x^{\text{ref}} \|^2 + \frac{\urho}{2} \|\ux - \un \|^2 \bigg].
\end{align}
The outgoing weights for this minimizer are always set to $\rho_0$.

Again by direct substitution we see that,
\begin{equation} \label{eq:minimizer_rvo_cost}
g^{\text{VO cost}}(\un,\urho,x^{\text{ref}},c) = g^{\text{cost}}\left(x^{\text{ref}},\un,+\infty,\urho,c \right),
\end{equation}
where in $g^{\text{cost}}$ we are using the energy minimizer for the global planning problem.

%
%

\section{Mechanical analog}
\label{app:mechanical_analog}

In this section we explain how to compute
the minimum energy configuration of the
springs-slab system described in
Section \ref{app:agent_obstacle_minimizer}.
Basically, it reduces to computing the minimum of a  one-dimensional real function over a closed interval.

Given $\un$ and $\on$, two main scenarios need to be
considered.
\begin{enumerate}
\item If $\un,\on \notin \mathcal{R}$ and $[\un\,\on] \cap \mathcal{R} = \emptyset$, i.e. the segment from $\un$ to $\on$ does not intersect $\mathcal{R}$, then $\ux = \un$ and $\ox = \on$.\\
\item Otherwise, because there might be multiple local minima, i.e. multiple stable static configurations, we need to compare the energy of the following two configurations and return the one with minimum energy.
\begin{enumerate}
\item The slab is
tangent to $\mathcal{R}$, for example as in the blue
configuration of Figure \ref{fig:mechanical_region}-right.\label{case:tagent}
\item  One of the springs is fully compressed and exactly one end of the slab is touching the boundary of $\mathcal{R}$, for example as in the red configuration of Figure \ref{fig:mechanical_region}-right.\label{case:not_tagent}
\end{enumerate}
\end{enumerate}

Let us compute the energy for Scenario \ref{case:tagent}. The slab can be tangent to $\mathcal{R}$ in many
different ways. However, the arrangement must always satisfy two properties. First, the point of contact, $p$, between the slab and $\mathcal{R}$ touches either the boundary of the semi-circle centered at $x_L$ or the boundary of the semi-circle centered at $x_R$. Second, because extending/compressing the slab costs zero energy, the slab must be orthogonal to the line segment $[\un \, \ux]$ and to the line segment $[\on \, \ox]$.

\begin{figure}[h!]
\begin{center}
\includegraphics[width=0.315\textwidth]{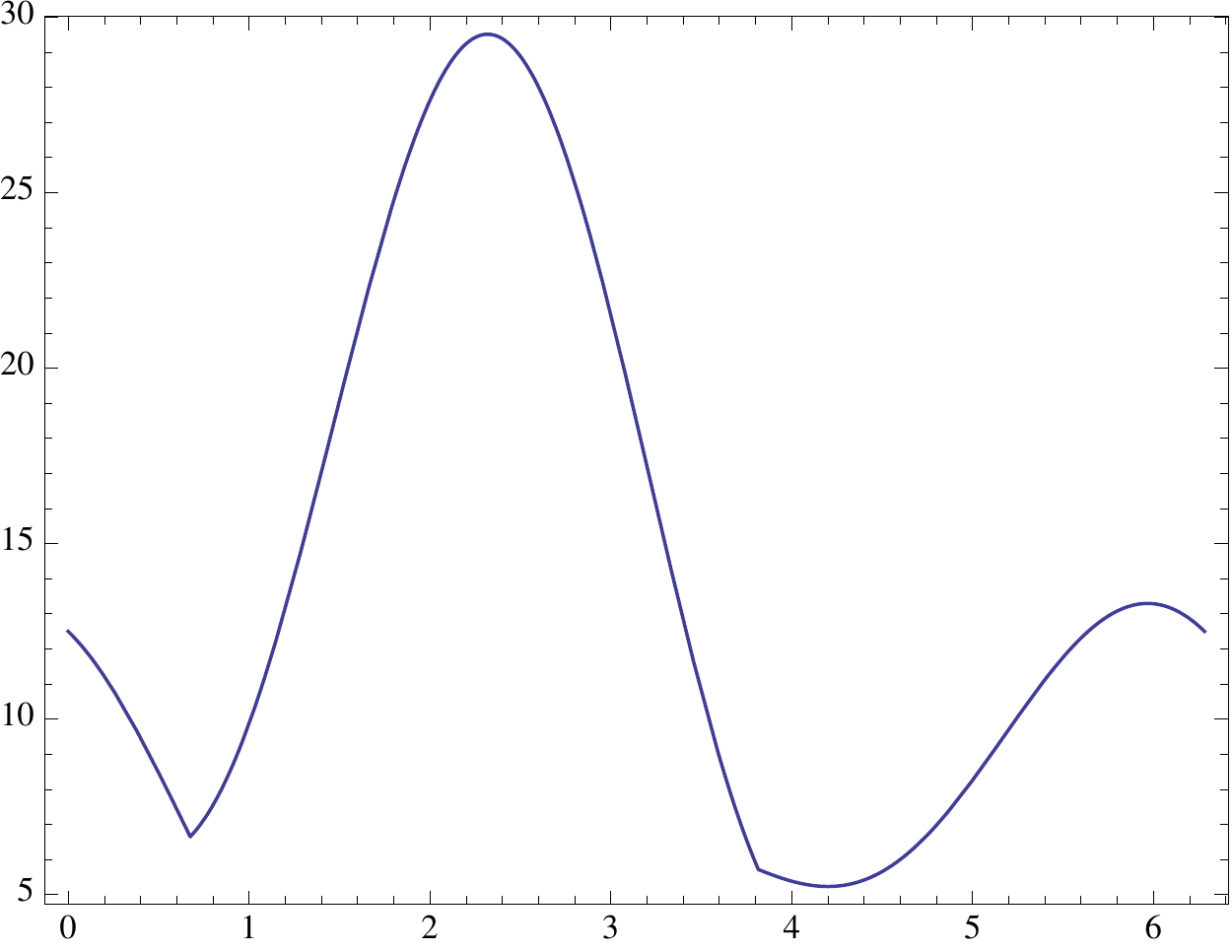}
\;\;\;\;\;\;\;\;\;\;\;\;\;\;\;\;\;\;\;\;\;\;\;\;\;\;\;\;
\includegraphics[width=0.315\textwidth]{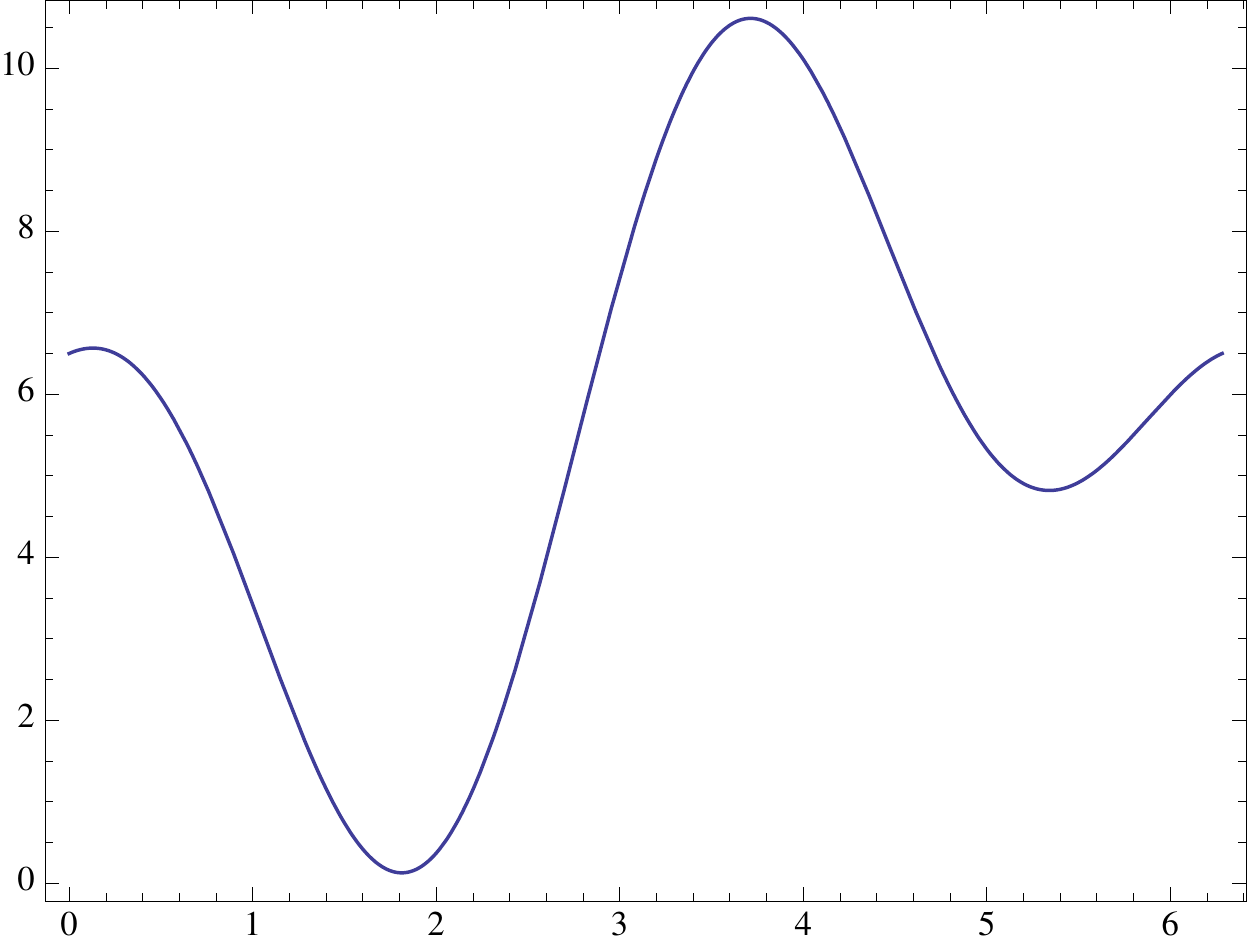}
\put(-60,-7){\tiny $\theta$}
\put(-100,50){\tiny $E_{\text{tangent}}(\theta)$}
\put(-271,-7){\tiny $\theta$}
\put(-300,50){\tiny $E_{\text{tangent}}(\theta)$}
\caption{Typical behavior of $E_{\text{tangent}}(\theta)$ for the agent-obstacle minimizer (left) and for the agent-agent minimizer (righ).}
\label{fig:one_dime_function}
\end{center}
\end{figure}

The first observation allows us to express $p$ using the map $P(\theta) : [0 , 2 \pi] \mapsto \reals^2$ between the direction of the slab and the point of contact at boundary of the semi-circles,
\begin{equation}
P(\theta) = \begin{cases}
x_R + r \hat{n}(\theta) & \langle x_R - x_L, \hat{
n}(\theta) \rangle  \geq \langle  x_L - x_R, x_R\rangle \\
x_L + r \hat{n}(\theta) & \text{otherwise}
\end{cases}
\end{equation}
where $\hat{n}(\theta) = \{\cos(\theta), \sin(\theta)\}$.
Specifically, there is a $\theta^0 \in [0, 2 \pi]$ such that $p = P(\theta^0)$.
The second observation tells us that $\ux = \un + \underline{\gamma} \hat{n}(\theta^0)$ and $\ox = \on + \overline{\gamma} \hat{n}(\theta^0)$ where $\underline{\gamma}$ and $\overline{\gamma}$ can be determined using the orthogonality conditions,
\begin{align}
&\langle \ux - P(\theta^0), \hat{n}(\theta^0) \rangle = 0 \Rightarrow \underline{\gamma} = \langle P(\theta^0) - \un , \hat{n}(\theta^0)\rangle\\
& \langle \ox - P(\theta^0), \hat{n}(\theta^0) \rangle = 0 \Rightarrow \overline{\gamma} = \langle P(\theta^0) - \on , \hat{n}(\theta^0)\rangle.
\end{align}
Therefore, the minimum energy configuration over all tangent configurations, which is fully determined by $\theta^0$, must satisfy
\begin{align}\label{eq:one_dime_prob_mech_analog}
\theta^0 &\in \arg \min_{\theta \in [0, 2 \pi]} E_{\text{tangent}}(\theta) \;\;\;\;\; \text{ where,}\\
E_{\text{tangent}}(\theta)  &= \frac{\urho}{2} (\langle P(\theta) - \un , \hat{n}(\theta)\rangle)^2  + \frac{\orho}{2} ( \langle P(\theta) - \on , \hat{n}(\theta)\rangle)^2 \label{eq:eq:one_dime_prob_mech_analog_function}.
\end{align}
Problem \eqref{eq:one_dime_prob_mech_analog} involves minimizing the one-dimensional function 
 \eqref{eq:eq:one_dime_prob_mech_analog_function}.
Figure \ref{fig:one_dime_function}-left shows the typical behavior of $E_{\text{tangent}}(\theta)$. It is non-differentiable in at most 2 points.
When the agent-obstacle minimizer is used to solve
the agent-agent minimizer, the function becomes
smooth and has second derivative throughout all its domain, see Figure \ref{fig:one_dime_function}-right. In the numerical results of Section \ref{sec:num}, our implementation of the agent-agent minimizer uses Newton's method to solve \eqref{eq:one_dime_prob_mech_analog}. To find the global minimum, we apply Newton's method starting from four equally-space points in $[0, 2\pi]$. To produce the video accompanying this appendix, our implementation of the agent-obstacle minimizer solves \eqref{eq:one_dime_prob_mech_analog} by scanning points in $[0, 2\pi]$ with a step size of $2 \pi /1000$. In this case, it is obvious there is  room for improvement in speed and accuracy by choosing smarter ways in which to solve \eqref{eq:one_dime_prob_mech_analog}.

To compute the energy for Scenario \ref{case:not_tagent}, we need to determine which of the springs is fully contracted, or which side of the slab is touching $\mathcal{R}$.
If $\un \in \mathcal{R}$ and $\on \notin \mathcal{R}$ then $\ox = \on$ and $\ux$ is the point in the boundary of $\mathcal{R}$ closest to $\un$ such that $[\ux \,\ox]$ does not intersect $\mathcal{R}$. Since the boundary of
$\mathcal{R}$ is formed by parts of the boundary of two circles and of two lines, this closest point can be computed in closed form. If $\on \in \mathcal{R}$ and $\un \notin \mathcal{R}$ the situation is the opposite.
If $\un,\on \in \mathcal{R}$, then we know we cannot be in Scenario \ref{case:not_tagent}. Finally, if $\un,\on \notin \mathcal{R}$, we compute the energy assuming that $\ux = \un$ and then assuming that $\ox = \on$ and take the configuration with smallest energy between them.

%
%

\section{Comment on our algorithm} 
\label{app:more_on_the_algorithm}
 
Note that our algorithm does \emph{not} possess anytime guarantees and, if stopped earlier, the trajectories might have collisions. However, if stopped early, a suboptimal set of non-colliding trajectories can be found at very low computational cost by using our algorithm to solve the feasibility problem in \eqref{eq:basic_opt_prob} starting from the state of the algorithm at stop time. In addition, although dynamic/static obstacles can be seamlessly integrated into our framework, a solution must be recomputed (as is the case with A$^*$ or RRT$^*$) if their trajectories/positions change unexpectedly. This being said, in our algorithm, if a new piece of information is received, the previous solution can be used as the initial guess, potentially decreasing convergence time. Note that in some scenarios, a low-cost local-planning approach, such as the one presented in Section \ref{sec:co}, can be beneficial.

%
%

\section{Comment on the scaling of convergence time with $p$} 
\label{app:more_on_the_convergence_time}

Based on \cite{nate13}, we think that in non-adversarial configurations, in contrast to CONF1 where dead-locks are likely, the scaling of convergence time with $p$ is not exponential. Our reasoning is based on a connection between trajectory planning and disk packing. For example, minimum-energy trajectory planning using piece-wise linear trajectories is related to, although not the same as, packing disks in multiple 2D layers, where the two matched disks between consecutive layers generate a larger cost when far away from each other. The numerical results in \cite{nate13} report that, for disk packing, the runtime of ADMM and TWA is no more than polynomial in the number of disks and we believe the runtime for trajectory planning for non-adversarial configurations has a similar complexity. We interpret the seemingly exponential curve of convergence time versus $p$ for $n$ = 8 in Figure\ref{fig:admm_vs_dynamic_admm_histogram_cores}-left as an atypical, adversarial scenario. By comparison, in Figure \ref{fig:feasible_vs_opt_energy_and_VO_speed}-left, which assumes randomly chosen initial and final points and also minimization of energy, the dashed-blue curve of runtime versus $p$ for $n$ = 8 does not appear to exhibit exponential growth.

\end{document}